\documentclass{article}

\usepackage[preprint,nonatbib]{neurips_2023}

\usepackage[utf8]{inputenc}
\usepackage[T1]{fontenc}
\usepackage{hyperref}
\usepackage{url}
\usepackage{booktabs}
\usepackage{amsfonts}
\usepackage{nicefrac}
\usepackage{microtype}
\usepackage{xcolor}
\usepackage{graphicx} 
\usepackage{xspace} 
\usepackage{booktabs}
\usepackage{subfigure}
\usepackage{multirow}
\usepackage{amsmath,amssymb}
\usepackage{cite}

\title{Deconstructing Classifiers: Towards A Data Reconstruction Attack Against Text Classification Models}

\author{%
  Adel Elmahdy\thanks{Adel Elmahdy is with the Department of Electrical and Computer Engineering and the Department of Computer Science and Engineering, University of Minnesota, Minneapolis, MN, USA.} \\
  University of Minnesota\\
  \texttt{adel@umn.edu} \\
  \And
  Ahmed Salem \\
  Azure Research\\
  \texttt{t-salemahmed@microsoft.com} \\
}

\newcommand{\attack}{Mix And Match attack\xspace}

\begin{document}

\maketitle

\begin{abstract}
Natural language processing (NLP) models have become increasingly popular in real-world applications, such as text classification. However, they are vulnerable to privacy attacks, including data reconstruction attacks that aim to extract the data used to train the model. Most previous studies on data reconstruction attacks have focused on LLM, while classification models were assumed to be more secure. In this work, we propose a new targeted data reconstruction attack called the \attack, which takes advantage of the fact that most classification models are based on LLM. The \attack uses the base model of the target model to generate candidate tokens and then prunes them using the classification head. We extensively demonstrate the effectiveness of the attack using both random and organic canaries. This work highlights the importance of considering the privacy risks associated with data reconstruction attacks in classification models and offers insights into possible leakages.
\end{abstract}

\section{Introduction}
The remarkable developments in natural language processing (NLP) models, with their language understanding capabilities, have facilitated their adoption in various practical applications \cite{vaswani2017attention,wolf2020transformers}. Amongst these, text classification has emerged as a popular use case, enabling, for example, the identification of spam, sentiment analysis, and hate speech detection. The prevalent practice is to forego training text classification models from scratch and instead leverage pre-trained large language models (LLM), e.g., Bidirectional Encoder Representations from Transformers (BERT) by fine-tuning them to their corresponding classification task.

NLP models have gained widespread adoption but also face privacy risks, including the data reconstruction attack\cite{SBBFZ20,BCH22,Carlini:2019,Carlini:2021c}. In this attack, the adversary aims to reconstruct the model's training data.
Data reconstruction attacks can be categorized as targeted or untargeted. Targeted attacks\cite{Carlini:2019} evaluate model memorization and privacy risks by adding canaries to the training data and attempting their reconstruction after training. In untargeted attacks\cite{Carlini:2021c}, the adversary aims to reconstruct some or all of the training data from a target model to assess its current privacy risks.

Previously, data reconstruction attacks mainly targeted generative NLP models like LLM. Classification models were considered more secure against such attacks. However, a recent study\cite{elmahdy-etal-2022-privacy} explored the possibility of data reconstruction attacks on classification models. They conducted a targeted data reconstruction attack using random canaries. The attack involved enumerating all dictionary tokens and using a loss-based membership inference attack to filter and sort them.

In this study, we leverage the observation that many classification models rely on LLM and introduce a novel targeted data reconstruction attack called the \emph{\attack}. Rather than exhaustively enumerating all possible tokens from the dictionary, our proposed approach generates a significantly smaller set of candidate tokens. Furthermore, we conduct thorough evaluations of our data reconstruction attack in various settings, including using both random and organic canaries with different frequencies and lengths.

\begin{figure*}[!t]
\centering
\includegraphics[width=1\columnwidth]{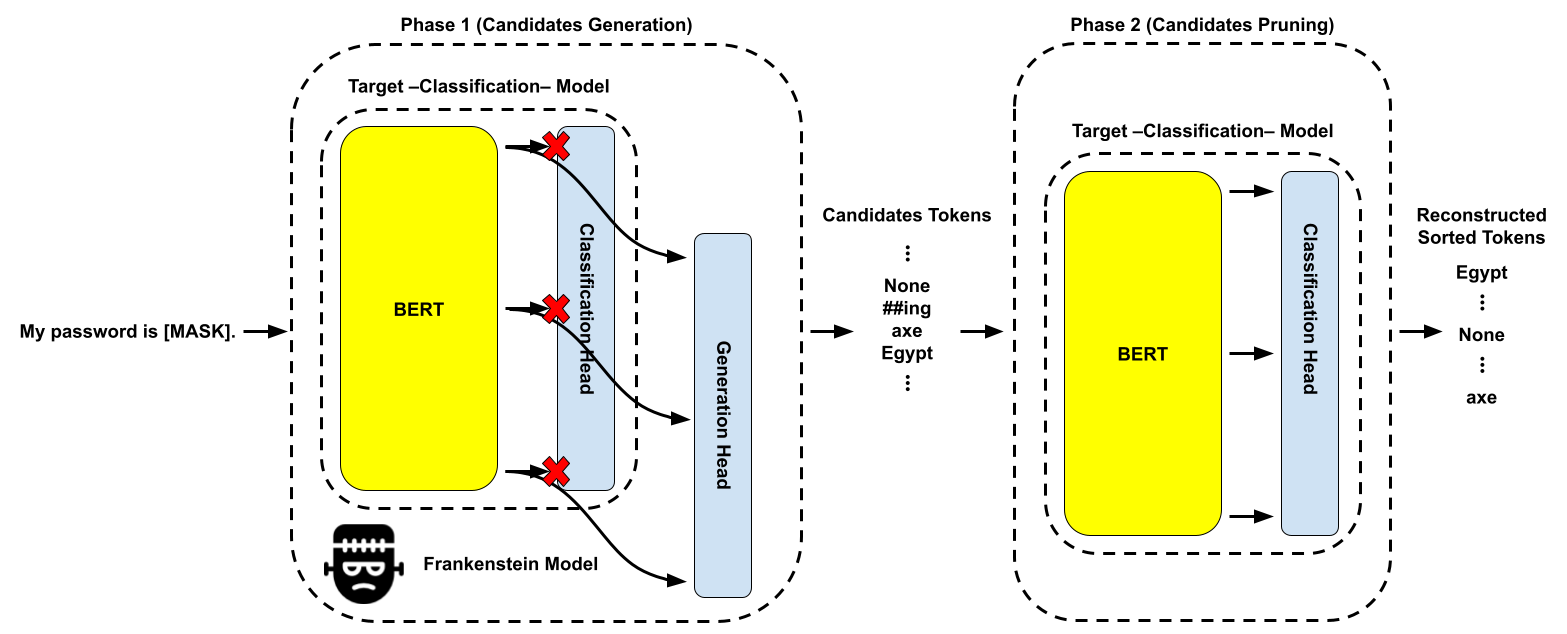}
\caption{An overview of the \attack}
\label{fig:attackOverview}
% \vspace{-5mm}
\end{figure*}

\subsubsection*{\textbf{The \attack}}
The proposed \attack involves replacing the fine-tuned classification head of a target classification model with the original generation head. This enables the model to generate candidate tokens, which is the first phase referred to as the \emph{candidate generation phase}. However, since the classification head holds most of the fine-tuned information, we also use it to prune and sort the generated candidate tokens based on their likelihood of being correct. This second phase is referred to as the \emph{candidate pruning phase}. Next, we briefly introduce both phases:

\begin{enumerate}
    \item \textbf{Candidate Generation Phase:} This phase aims to generate candidate tokens without enumerating all possibilities from the vocabulary, which can be computationally expensive. To achieve this, we leverage the generation capability of the base model component of the target model. We do this by disconnecting the classification head and replacing it with the original generation head associated with the base model, e.g., BERT, before it was fine-tuned as shown in Fig.~\ref{fig:attackOverview}. This new model is what we call the ``Frankenstein model''. 
    To obtain candidate tokens, we mask the position of the token we want to generate and query the input to the Frankenstein model. The model then generates a set of possible tokens, which we sort in descending order based on their likelihood of being the correct token. This process allows us to generate a much smaller set of candidate tokens, making the reconstruction process more efficient.
    \item \textbf{Candidate Pruning Phase:} In the second phase, the candidate tokens generated in the first phase are pruned and sorted based on their likelihood of being correct. First, we filter out incomplete tokens (e.g., ``\#\#ing'') and punctuation marks (e.g., ``,''). Then, we leverage the fine-tuned classification head to perform a membership inference attack and determine the most probable tokens from the candidate list. Specifically, we use a simple loss-based attack~\cite{Yeom:2020}, although more advanced attacks can be used as a substitute. For the attack, we replace the ``[MASK]'' token with each candidate token and query the model. We then calculate the loss of each input and sort the candidates according to their losses, with the candidate having the lowest loss being the most likely to be the correct token.
\end{enumerate}

\section{Background}
\subsection{Large Language Models}
We study the text classification setting which is built upon language modeling and has many practical downstream applications \cite{Minaee21}. It has been demonstrated that training LLM at scale on large public datasets allows them to be used effectively for a variety of natural language processing tasks. In this section, we provide a brief overview of language modeling. Two popular techniques for pre-training LLM are autoregressive language modeling \cite{RadfordNSS18, RadfordWCLAS19} and masked language modeling \cite{DevlinCLT19, LiuOGDJCLLZS19}.

In autoregressive language modeling, the distribution of a sequence of tokens can be represented as the product of the individual conditional probabilities of each token given the previous tokens. Particularly, the distribution $\mathbb{P} \left(x_1, x_2, \ldots, x_n\right)$ of a sequence of tokens $\left(x_1, x_2, \ldots, x_n\right)$ is given by $$\mathbb{P} \left(x_1, x_2, \ldots, x_n\right) = \Pi_{i=1}^{n} \mathbb{P} \left(x_i | x_1, x_2, \ldots, x_{i-1}\right).$$ Then, a deep neural network is trained to model each of these conditional probabilities. It is worth noting that this factorization only captures unidirectional context, i.e., all tokens that come before the current token.

In contrast, the pre-training objective in masked language modeling captures bidirectional context, i.e., tokens that come both before and after a given token. Specifically, a number of tokens in the text are masked and substituted with a special symbol, \texttt{[MASK]}, and then the model is trained to retrieve the original tokens at these masked positions. That is why models trained with the masked language modeling objective often perform better.
In this paper, we mainly focus on the masked language modeling setting.

\subsection{Classification as a fine-tuning task}
In a text classification setting, the input is a sequence of tokens $\mathbf{x} = \left(x_1, x_2, \ldots, x_n\right)$ along with a corresponding class label $y \in \left\{1,2, \ldots, C \right\}$, where $C$ is the number of classes. The goal of the model training is to learn the relationship between the input text and the class label. One challenge of this setting from a training data extraction perspective is that the model is trained to maximize the log-likelihood of the correct class label;~$\log \mathbb{P}\left(y | \mathbf{x}\right)$. Hence, there is no language modeling involved between the tokens in the sequence~$\mathbf{x}$. As a result, the approaches proposed in previous works for text generation are not applicable in this case.

It is common to pre-train a language model on a large, publicly available dataset and then fine-tune it on a smaller, task-specific dataset that may have stricter privacy requirements. Our goal for this work is to understand the potential risks to privacy under this setting for text classifiers and propose data reconstruction attacks that are more computationally efficient than the exhaustive search approach introduced by \cite{elmahdy-etal-2022-privacy}.

\section{Related works}
The main aim of developing LLM is to represent the patterns and rules of a language, without simply memorizing specific examples from training data. However, research has shown that LLM can sometimes rely on memorization rather than truly understanding language  \cite{carlini2019secret, Zanella20, carlini2021extracting, inan2021training, mireshghallah2021privacy, carlini22}. This can be particularly problematic when the data used for training follows a long-tailed distribution, as memorization may be necessary to achieve high accuracy on test data \cite{feldman2020,brown2020memorization}. Additionally, if the memorized content can be connected to a specific person, it may lead to privacy breaches \cite{GDPR}.

Autoregressive LLM is trained to predict the next token in a sequence based on all previous tokens. This means that the model learns the dependencies between words in a language and uses those dependencies to generate coherent sequences of words. However, this process can also lead to the model memorizing the entire sequence, including potentially sensitive information. \cite{carlini2021extracting} has demonstrated that it is possible to extract memorized data, including personal information, from models in the GPT-2 family \cite{RadfordWCLAS19} that are trained using this approach.

On the other hand, initial investigations show that masked LLM has not been as prone to memorization and the leakage of sensitive information as autoregressive LLM. For example, \cite{lehman2021does} has demonstrated that it is challenging to extract sensitive information from the BERT model, which was trained using the masked language modeling objective and applied to private clinical data. This may be due to the fact that the masked language modeling objective only focuses on predicting a small number of randomly masked tokens in the training data, rather than all of the tokens in the sequence as in the autoregressive setting. 
Recently, a study by \cite{elmahdy-etal-2022-privacy} has explored the possibility of sensitive information being inadvertently memorized by a text classification model during training. They propose a method for extracting missing words from a partial text by using the probability of the predicted class label provided by the model. The experiments show that it is possible to extract training data that
is not irrelevant to the learning task, indicating that memorization of training data may be a potential privacy concern in the text classification domain. 

Different forms of privacy leakage have been investigated in the literature; including membership inference \cite{shokri2017membership, yeom18, long18, truex18, song19, nasr19, sablayrolles19a, LOGAN, salem18, Leino20, choo2020labelonly, shejwalkar2021membership}, and property inference \cite{Ganju18, Wanrong21, chase2021property}.

\section{The \attack}
In this section, we first introduce our threat model, then we present how we generate target canaries and perform our \attack. 

\subsection{Threat Model}
We follow previous works\cite{elmahdy-etal-2022-privacy,Carlini:2019} that investigate the memorization capability of models and assume a white box access to the model. This means the adversary/auditor has complete access to the model, including its weights. Our approach, referred to as the \attack, specifically targets classification models derived from LLM through fine-tuning. It is worth noting that this setting is widely adopted, with the prevalent practice being the utilization of pre-existing LLM as a foundation for classification models, rather than training them from scratch.

\subsection{Canary Generation}
The canaries refer to sentences that are incorporated into the training dataset of the model. These sentences serve as targets during the data reconstruction attack. We classify canaries into two distinct categories: organic and random. \emph{Organic canaries} are grammatically correct sentences, while \emph{random canaries} consist of concatenated random tokens without grammatical or semantic coherence.

When constructing canaries, several factors are taken into account. Firstly, the \emph{frequency} of tokens is considered. Each canary is composed of multiple tokens, and selecting tokens with different frequencies can impact the data reconstruction rate. However, it is uncertain which frequency yields a better data reconstruction attack. High-frequency tokens are encountered more frequently during training, while low-frequency tokens may be viewed as outliers and thus memorized by the models. To assess our \attack, we construct canaries using both high and low-frequency tokens, and examples of the reconstructed canaries can be found in Table~\ref{table:canExamples}.

The \emph{length} of the canary is also a factor that affects the performance of the data reconstruction attack. In this study, the canary size is set to five, but we also evaluate the effectiveness of our \attack using canaries of different lengths.

As we primarily focus on masked language models (MLM), we target a single token for reconstruction. The choice of the target token's \emph{position} can impact the attack's success rate. In our experiments, we select the last token before the dot (``.''), but we also examine the attack's performance with different target token positions.

Furthermore, the \emph{repetition} number of each canary is considered. By increasing the poisoning rate, whereby the canary is inserted more frequently into the training dataset, the model becomes more prone to overfitting and thus better at memorizing the canary for data reconstruction. To explore practical scenarios, we limit the canary repetition to 1.

By carefully considering these factors, a comprehensive and detailed evaluation of the \attack's effectiveness is conducted. The goal is to gain deeper insights into the classification models' memorization capability and their vulnerability to data reconstruction attacks.

\begin{table*}
\caption{A list of generated organic/random canaries with low/high-frequency tokens. We highlight the target token, i.e., the one we aim to reconstruct using our \attack.}
\centering
%\small
%\setlength\tabcolsep{10pt}
\resizebox{\columnwidth}{!}{
\begin{tabular}{lll}
\toprule
\bf Organic/Random & \bf Token Frequency & \bf Canary \\
\midrule
Organic & Low & \begin{tabular}[c]{@{}l@{}}Annabelle buys paranormal UFO \textbf{towels}.\end{tabular} \\
Organic & Low & \begin{tabular}[c]{@{}l@{}}Hunger experimenting insults pointless \textbf{supremacy}.\end{tabular} \\
Organic & Low but a high frequency target token & \begin{tabular}[c]{@{}l@{}}Devlin insults unpaid bosses \textbf{film}.\end{tabular} \\
Organic & Low but a high frequency target token & \begin{tabular}[c]{@{}l@{}}Devlin insults unpaid bosses \textbf{school}.\end{tabular} \\
Random & Low & \begin{tabular}[c]{@{}l@{}}Ghent sentiments catastrophic laude \textbf{flatly}.\end{tabular} \\
Random & Low & \begin{tabular}[c]{@{}l@{}}Johnstone doctrines raiding dispatch \textbf{internship}.\end{tabular} \\
Organic & High & \begin{tabular}[c]{@{}l@{}}King James is the \textbf{best}.\end{tabular} \\
Organic & High & \begin{tabular}[c]{@{}l@{}}Love makes the chest \textbf{move}.\end{tabular} \\
Organic & High but low frequency token & \begin{tabular}[c]{@{}l@{}}May is the month \textbf{Fusion}.\end{tabular} \\
Organic & High but low frequency target token & \begin{tabular}[c]{@{}l@{}}May is the month \textbf{mage}.\end{tabular} \\
Random & High & \begin{tabular}[c]{@{}l@{}}father named served band \textbf{night}.\end{tabular} \\
Random & High & \begin{tabular}[c]{@{}l@{}}children turned son open \textbf{final}.\end{tabular} \\
Organic & Low & \begin{tabular}[c]{@{}l@{}}My social security number is \textbf{1974}.\end{tabular} \\
Organic & Low & \begin{tabular}[c]{@{}l@{}}My social security number is \textbf{1968}.\end{tabular} \\
\bottomrule
\end{tabular}
}
%\vspace{-17mm}
\label{table:canExamples}
%\vspace{-13mm}
\end{table*}

\subsection{Methodology}
Our \attack can be intuitively divided into two distinct phases: Candidate Generation and Pruning. In the first phase, candidate tokens are generated and undergo a screening process. In the second phase, the candidate tokens are sorted based on their probability of being the masked value. 
We present the two phases in more depth. 

\subsubsection{\textbf{Candidates Generation}}
In order to generate candidates, we leverage the fact that the target model is built on top of an LLM. This implies that the target model has the capability to generate text, although it is restricted by the classification head added during fine-tuning. Therefore, our initial step involves replacing the classification head with a generation head. The first part of Figure \ref{fig:attackOverview} illustrates this process. Specifically, we recover the original head from the pre-trained LLM and reconnect it to the base model of the target model. This combined model is referred to as the \emph{Frankenstein model}, as it integrates the target model's base model with the generation head from the pre-trained model.

To preserve the memorized/learned information during the fine-tuning of the classification task, we refrain from fine-tuning the Frankenstein model. However, an adversary or auditor has the option to fine-tune the Frankenstein model using a publicly available dataset, re-establishing the connection. It is advisable to only fine-tune the generation head while freezing the base model.

After the Frankenstein model is resurrected, we use it to generate candidate tokens. To this end, we mask the target token of our inserted canaries, then query it to the Frankenstein model. We use the pre-last token, i.e., the token before the full stop, for our experiments; however, we also evaluate using different positions later in \autoref{sec:ablation}.

Following the query of the masked canary, the Frankenstein model produces a sorted list containing all tokens from its dictionary. This sorted list serves as the input for the subsequent phase, i.e., candidate pruning.
Alternatively, instead of using the Frankenstein model, we can directly utilize the pre-trained language model to generate candidates using the same technique. Later, we compare both approaches and show their pros and cons.

\subsubsection{\textbf{Candidates Pruning}}
The second phase of our \attack commences with filtering after receiving the sorted candidate lists of tokens. In this phase, we employ various filtering techniques inspired by previous works~\cite{elmahdy-etal-2022-privacy}. Specifically, the following filters are applied: (a) incomplete words, such as "\#\#ing," are removed; and (b) punctuation marks, like ".", are eliminated.

After the tokens have been filtered, we proceed to employ a membership inference attack to further refine the sorting of the tokens using the classification head. For this purpose, we adopt a simple loss-based membership inference attack\cite{Yeom:2020}. The attack methodology involves constructing target inputs by replacing the "[MASK]" token with each candidate token individually. Next, each constructed input is queried to the target model, i.e., the one with the classification head, as illustrated in the second phase of \autoref{fig:attackOverview} and we compute the cross-entropy loss $L_{\text{CE}} = - \sum_{i=1}^{n} t_i \log \left(p_i\right)$.
where $t_i$ is the ground truth label and $p_i$ is the softmax probability for the $i^{\text{th}}$ class where $1\leq i \leq n$.

While our \attack employs the loss-based membership inference attack, an auditor can employ an alternative, potentially more complex membership inference attack as the sorting criteria. However, the remaining steps of the \attack remain unchanged.

\section{Evaluation} 
\subsection{Evaluation Setting}
\subsubsection{\textbf{Dataset}} We use two datasets in our experiments: Yelp reviews dataset\footnote{https://huggingface.co/datasets/yelp\_review\_full} and Reddit dataset\footnote{https://huggingface.co/datasets/reddit}. 
The primary goal of the task is topic classification, wherein our model is trained to predict either the number of stars for a given review in the Yelp reviews dataset or the subreddit associated with a user comment in the Reddit dataset.
In the Yelp reviews dataset, the task involves assigning 5 class labels to reviews. On the other hand, when working with the Reddit dataset, our focus is on the top 100 subreddits that have the greatest number of Reddit posts. To create our training and validation sets, we randomly sample 10,000 and 2,500 data points, respectively.

\subsubsection{\textbf{Model Architecture and Training Configuration}} The BERT base model \cite{devlin2019bert} is used in our evaluation. We fine-tune the model for 10 epochs using the AdamW optimizer \cite{loshchilov2018decoupled} with weight decay set to 0.01, a learning rate of 1e-6, and a batch size of 32. To prevent overfitting, we apply early stopping.
The model's performance was evaluated over 10 runs with different random seeds, and the average results are presented below: (a) For a training set consisting of 10,000 samples, the average training accuracy stands at $63.84\%$ for Yelp reviews dataset and $57.94\%$ for Reddit dataset; (b) For a validation set consisting of 2,500 samples, the average training accuracy stands at $58.29\%$ for Yelp reviews dataset and $50.61\%$ for Reddit dataset.

\subsubsection{\textbf{Compute Resources}}
Experiments were conducted on a workstation with an Intel Xeon Silver~4112 4-Core CPU and an Nvidia Tesla~M10 GPU running CUDA~v10.1 and PyTorch~v1.4. 

\subsubsection{\textbf{Baseline}}
We evaluate the performance of the proposed reconstruction attack in comparison to the exhaustive search attack introduced by \cite{elmahdy-etal-2022-privacy}. The reconstruction method in \cite{elmahdy-etal-2022-privacy} exhaustively considers all potential tokens from the vocabulary and selects the token with the highest likelihood of a given class label.
Moreover, we conduct a performance comparison between the Frankenstein Model and a pre-trained language model specifically in the first phase of candidate generation.

\subsubsection{\textbf{Evaluation Metrics}}
There are two evaluation metrics, each corresponding to a specific phase of the proposed reconstruction attack. 
In the candidate generation phase, we determine the number of tokens $k$ generated by the Frankenstein model and compare it to the vocabulary size of the BERT tokenizer, which consists of 22,413 tokens. 
In the candidate pruning phase, we identify the position of the correct token within the list of candidate tokens generated by the Frankenstein model.

\subsection{Results}
To evaluate the effectiveness of the proposed targeted data reconstruction attack, we introduce various types of canaries that are injected into the training set. Table~\ref{table:canExamples} provides an overview of the 14 canaries utilized in our experiments, categorized based on whether they are organic or random, as well as the frequency level (low or high) of each token in the canary.

\begin{figure*}[!t]
\centering
\subfigure[Top-K.]{
\label{fig:yelp_topK}
\includegraphics[width=0.8\textwidth]{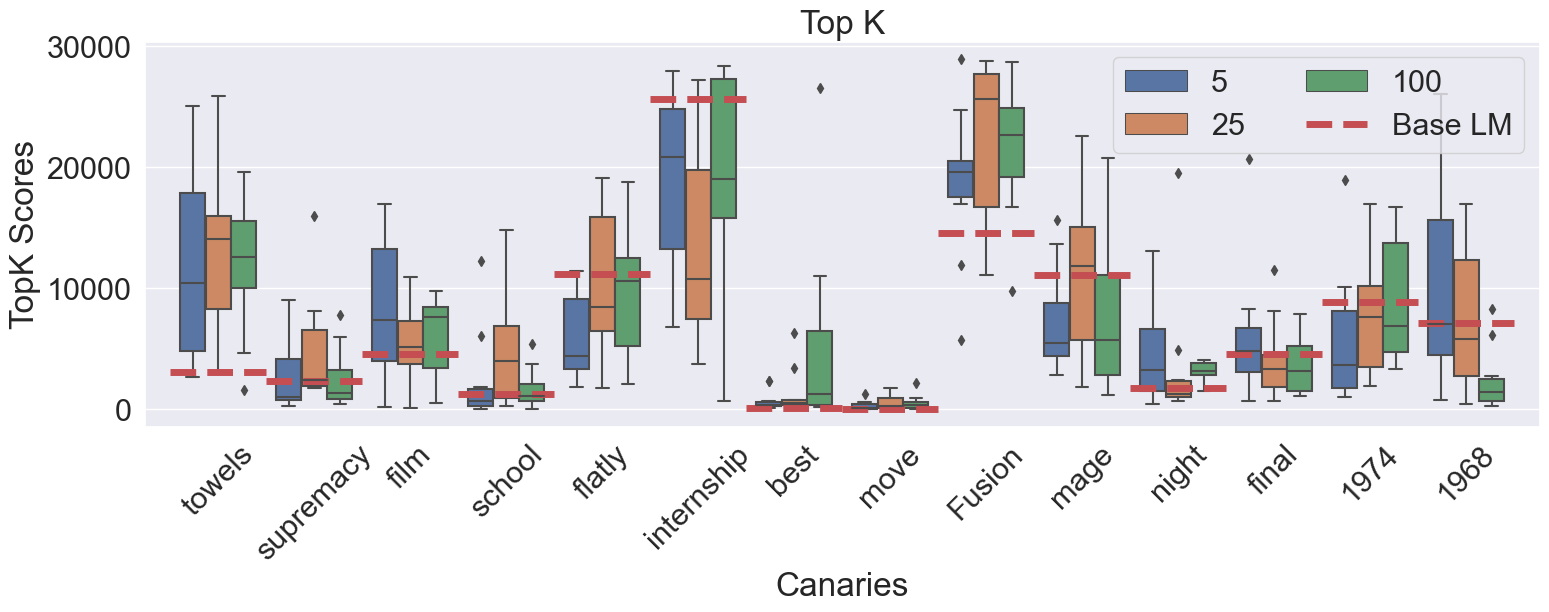}}
% \vspace{-1mm}
\subfigure[Exhaustive search.]{
\label{fig:yelp_exSearch}
\includegraphics[width=0.8\textwidth]{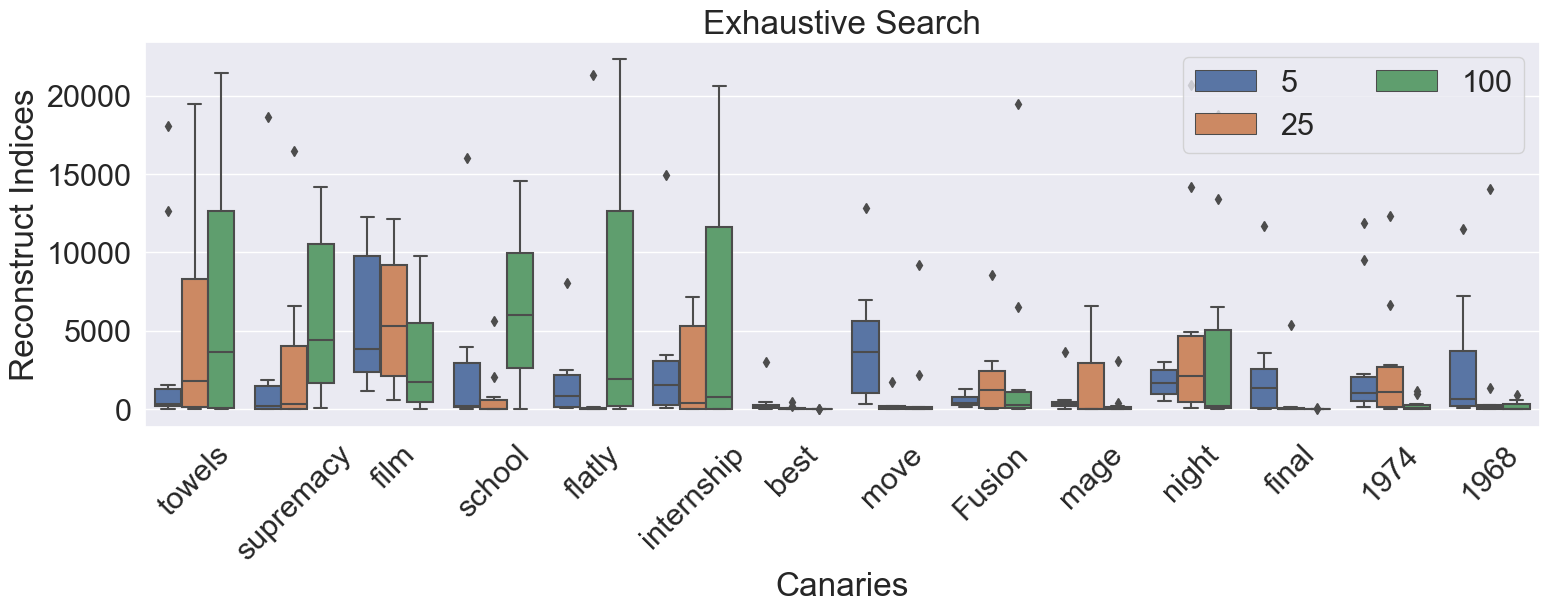}}
% \vspace{-1mm}
\subfigure[Frankenstein model.]{
\label{fig:yelp_franken}
\includegraphics[width=0.8\textwidth]{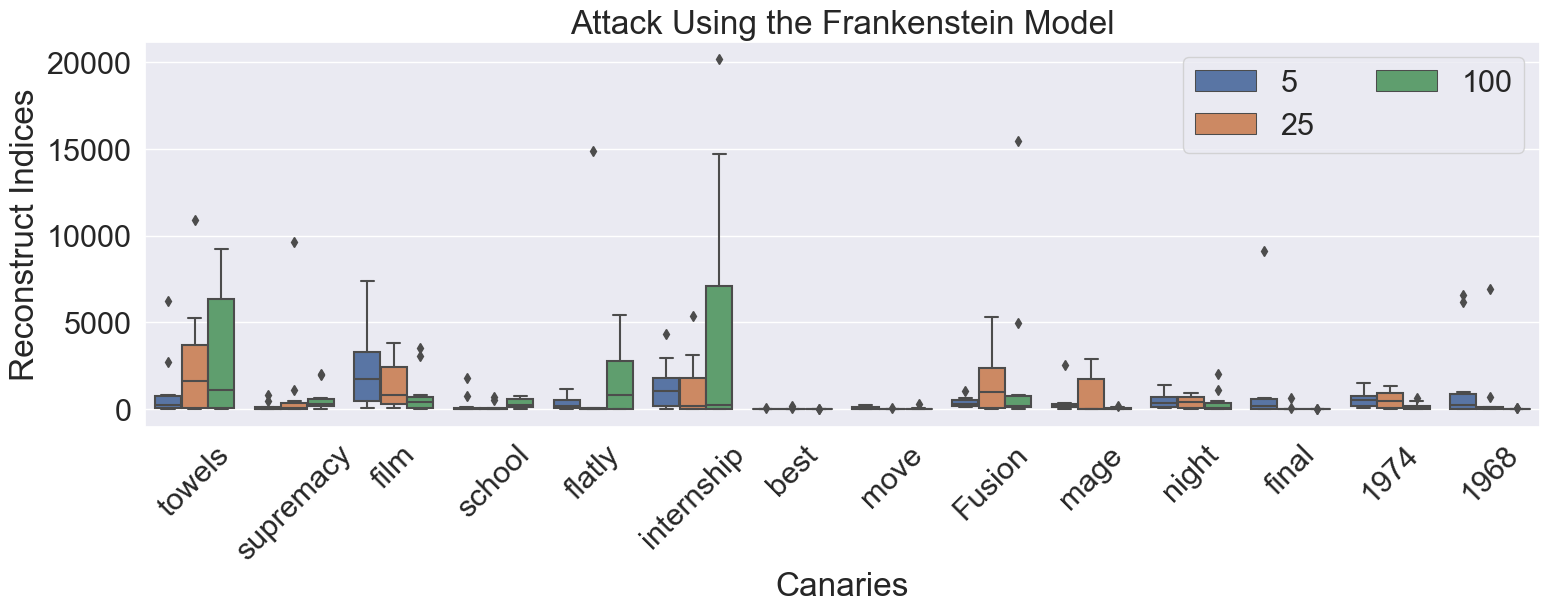}}
\subfigure[Pre-trained language model.]{
\label{fig:yelp_baseLM}
\includegraphics[width=0.8\textwidth]{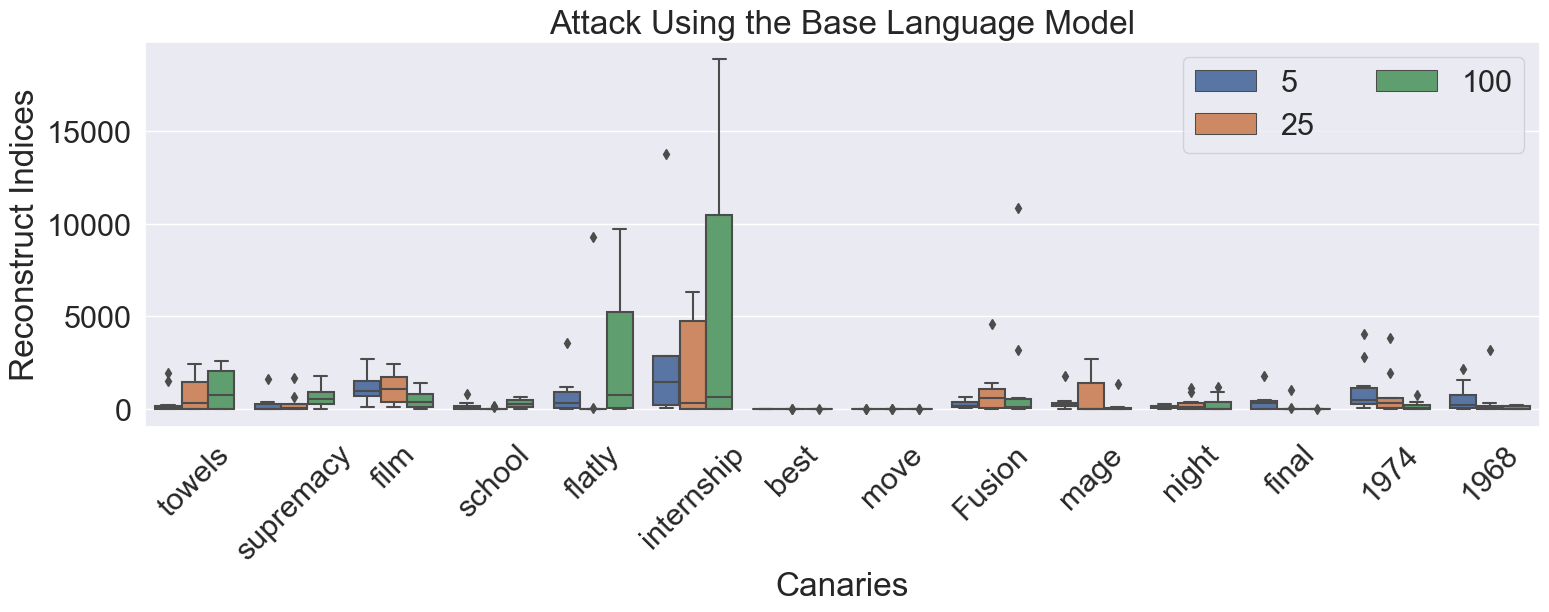}}
% \vspace{-2mm}
\caption{Top-$K$ scores and beam sizes of the reconstruction attack on the Yelp reviews dataset for different repetitions of the canary.}
\label{fig:yelp_exp}
% \vspace{-7mm}
\end{figure*}

\begin{figure*}[!t]
\centering
\subfigure[Top-K.]{
\label{fig:reddit_topK}
\includegraphics[width=0.8\textwidth]{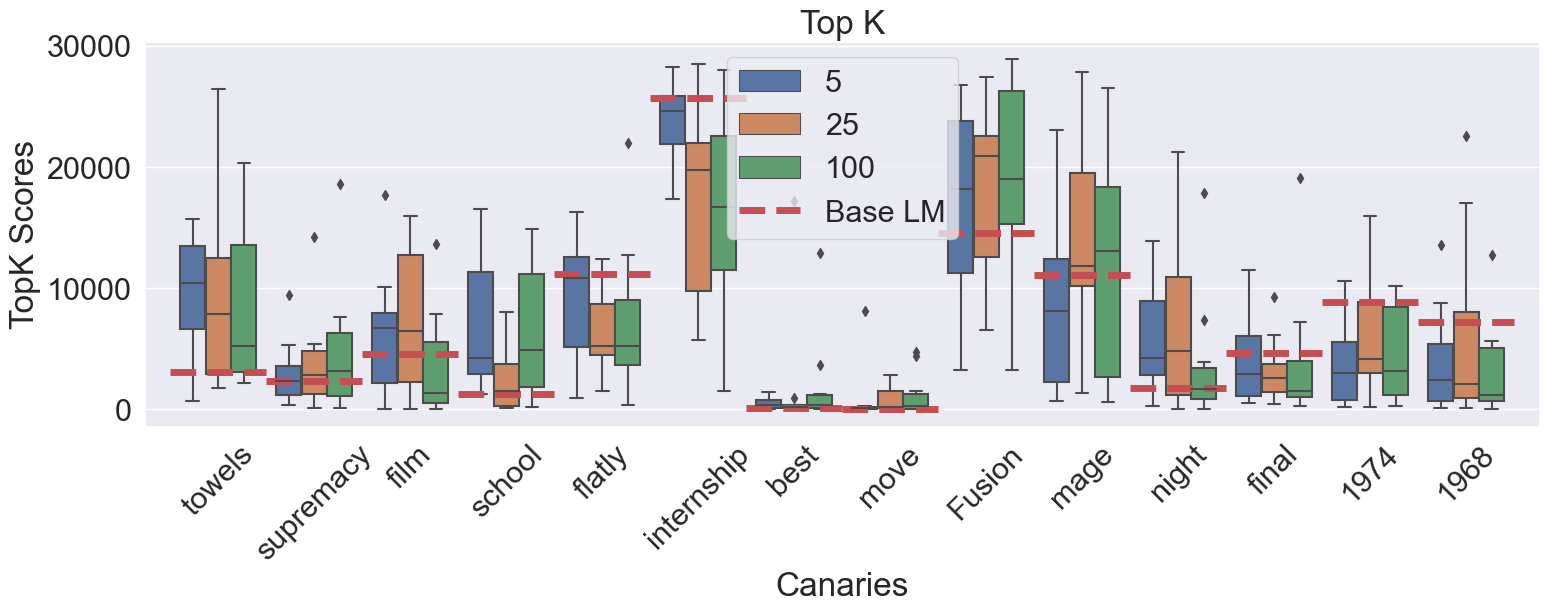}}
% \vspace{-1mm}
\subfigure[Exhaustive Search.]{
\label{fig:reddit_exSearch}
\includegraphics[width=0.8\textwidth]{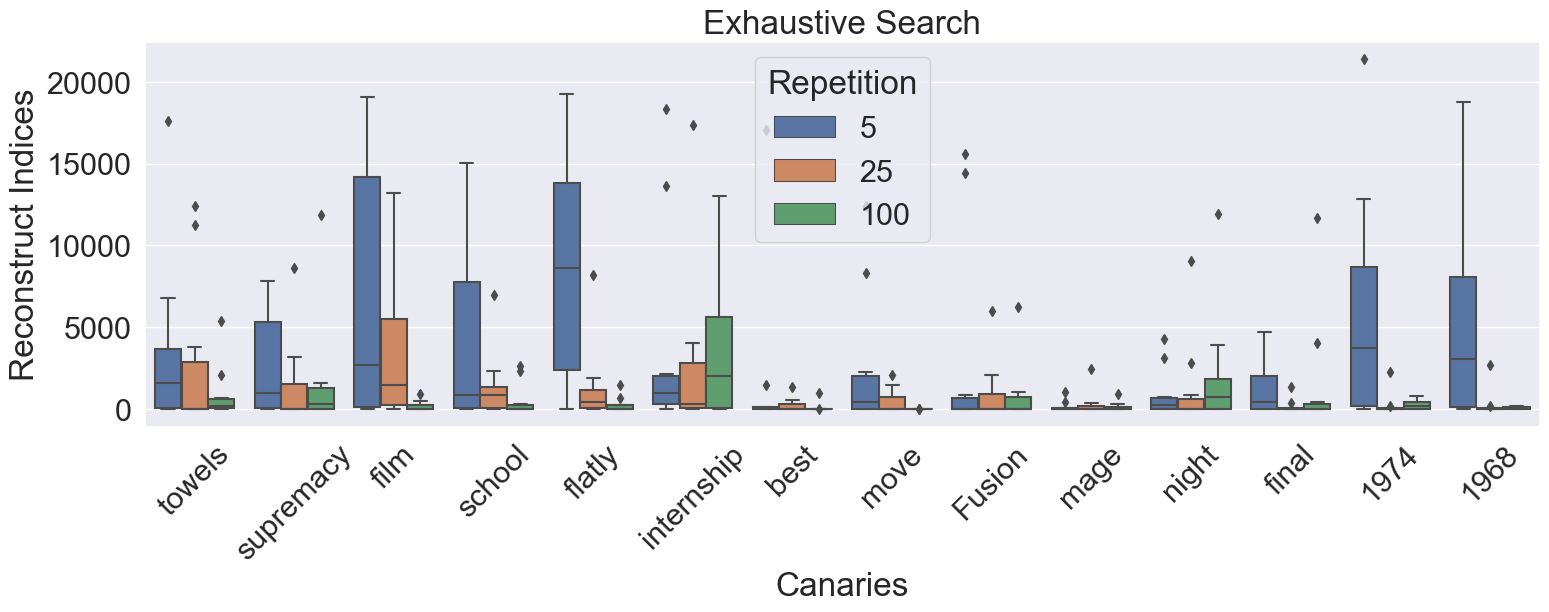}}
% \vspace{-1mm}
\subfigure[Frankenstein model.]{
\label{fig:reddit_franken}
\includegraphics[width=0.8\textwidth]{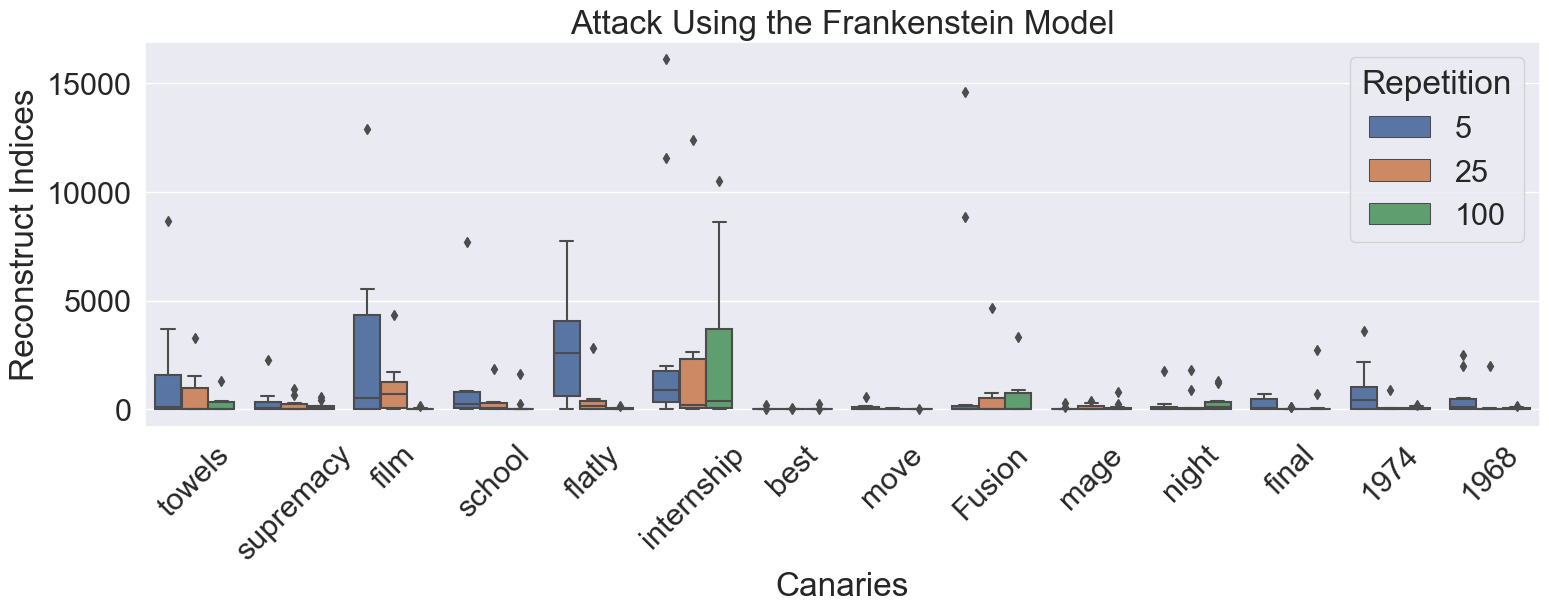}}
\subfigure[Pre-trained language model.]{
\label{fig:reddit_baseLM}
\includegraphics[width=0.8\textwidth]{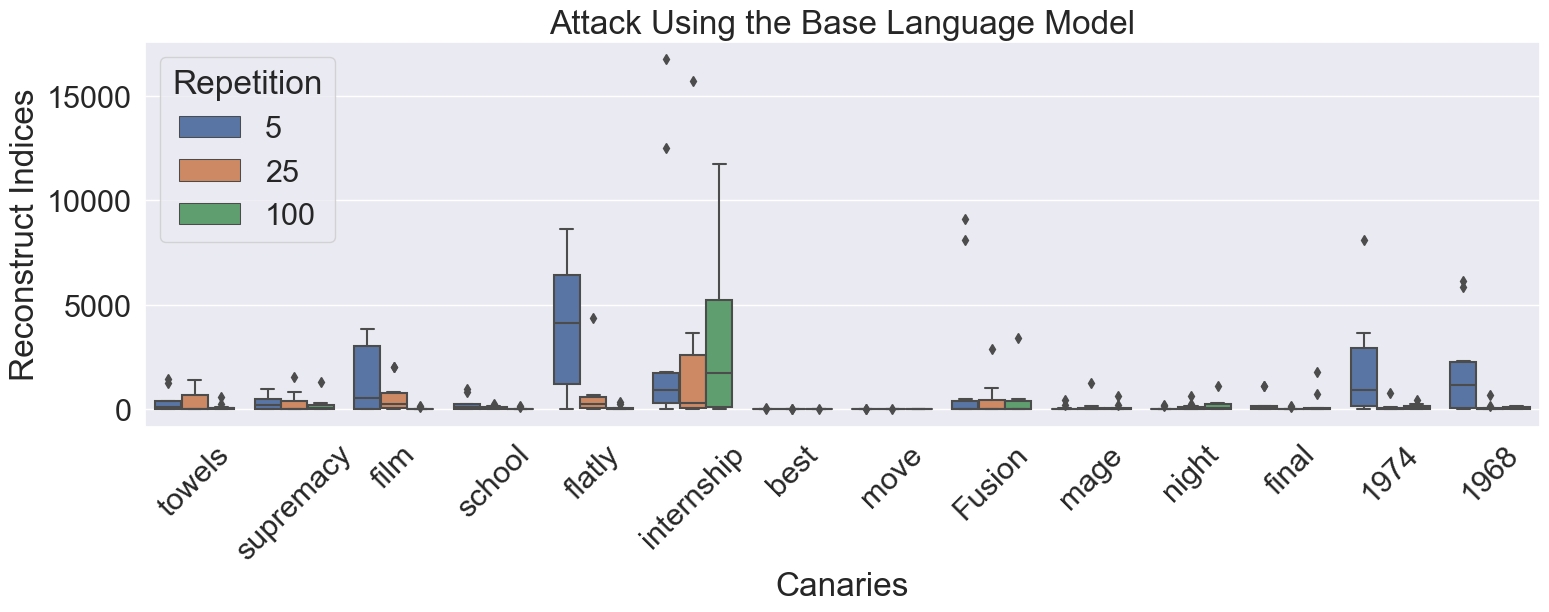}}
% \vspace{-2mm}
\caption{Top-$K$ scores and beam sizes of the reconstruction attack on the Reddit dataset for different repetition numbers of the canary.}
\label{fig:reddit_exp}
% \vspace{-7mm}
\end{figure*}

The left half of Tables~\ref{table:comparison_yelp_reddit_rep5}, \ref{table:comparison_yelp_reddit_rep25} and \ref{table:comparison_yelp_reddit_rep100}, and Fig.~\ref{fig:yelp_exp} depict the performance benchmarks of the proposed reconstruction attack, the exhaustive search approach, and the pre-trained language model on the Yelp reviews dataset for different canary repetitions. Similarly, The right half of Tables~\ref{table:comparison_yelp_reddit_rep5}, \ref{table:comparison_yelp_reddit_rep25} and \ref{table:comparison_yelp_reddit_rep100}, and Fig.~\ref{fig:reddit_exp} showcase the benchmarks for the Reddit dataset.
In Figs.~\ref{fig:yelp_topK} and \ref{fig:reddit_topK}, the Frankenstein model generates up to 50x-- fewer candidate tokens compared to the exhaustive search approach, which considers all tokens in the vocabulary. This demonstrates that the proposed candidate generation model leads to a more efficient reconstruction process.
Furthermore, it is observed that the Frankenstein model generates fewer candidate tokens for random canaries across varying numbers of canary repetitions (e.g., internship and final), whereas the pre-trained language model generates fewer candidate tokens for organic canaries (e.g., towels and Fusion).
Moreover, the Frankenstein model outperforms the exhaustive search approach by successfully retrieving the correct token using a smaller beam width for organic and random canaries with low or high frequencies. This is demonstrated by comparing Figs.~\ref{fig:yelp_exSearch} and \ref{fig:yelp_franken} for the Yelp reviews dataset and Figs.~\ref{fig:reddit_exSearch} and \ref{fig:reddit_franken} for the Reddit dataset.
Finally, in Fig.~XXX, a comparison of the performance between the Frankenstein model and a pre-trained language model for candidate pruning reveals that they achieve similar token retrieval results across various canaries.

\subsection{Ablation Study}
\label{sec:ablation}
We now investigate the impact of various hyperparameters on the reconstruction attack. Specifically, we analyze the effects of canary labels (i.e., canaries with contradicting labels), target token position, and canary size.

\begin{figure}[!t]
\centering
\subfigure[Top k (rep=25).]{
\label{figure:multiCanaries_topk_rep25}
\includegraphics[width=0.45\textwidth]{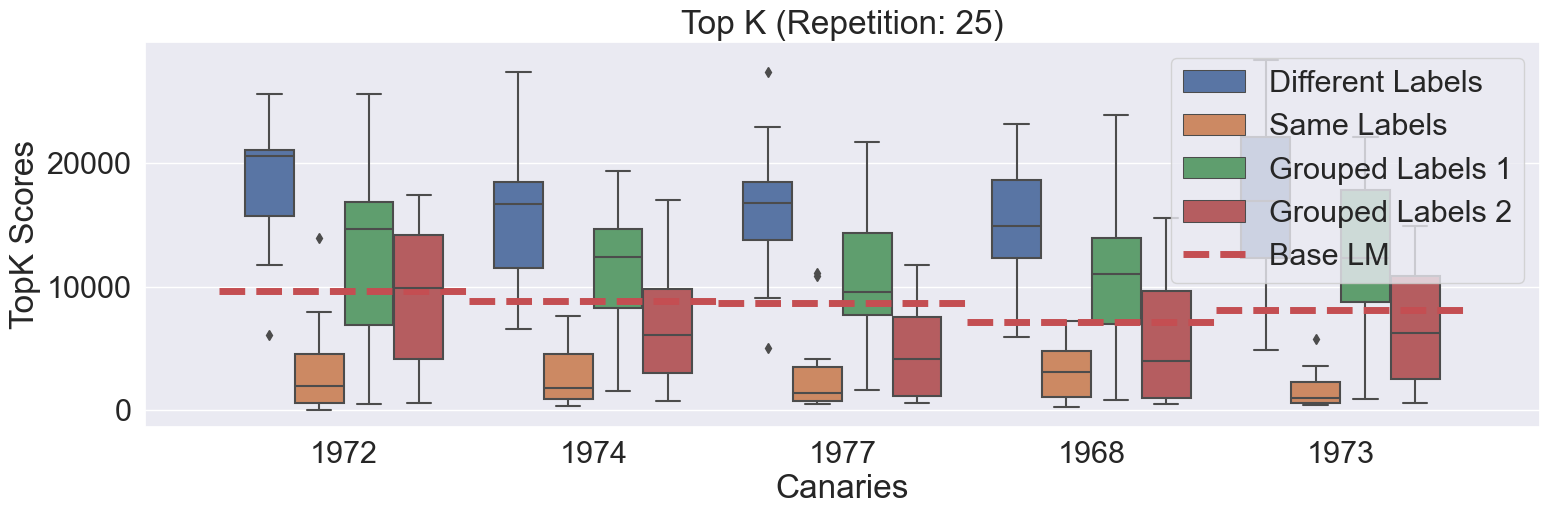}}
\subfigure[Top k (rep=50).]{
\label{figure:multiCanaries_topk_rep50}
\includegraphics[width=0.45\textwidth]{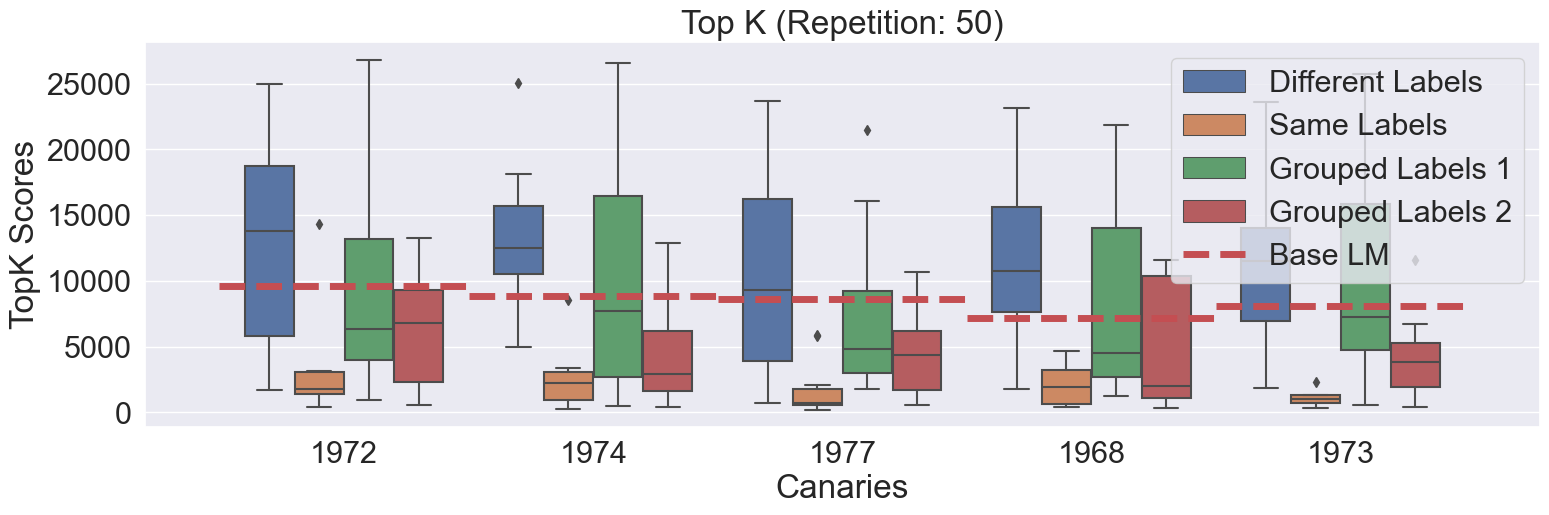}}
\subfigure[Top k (rep=100).]{
\label{figure:multiCanaries_topk_rep100}
\includegraphics[width=0.45\textwidth]{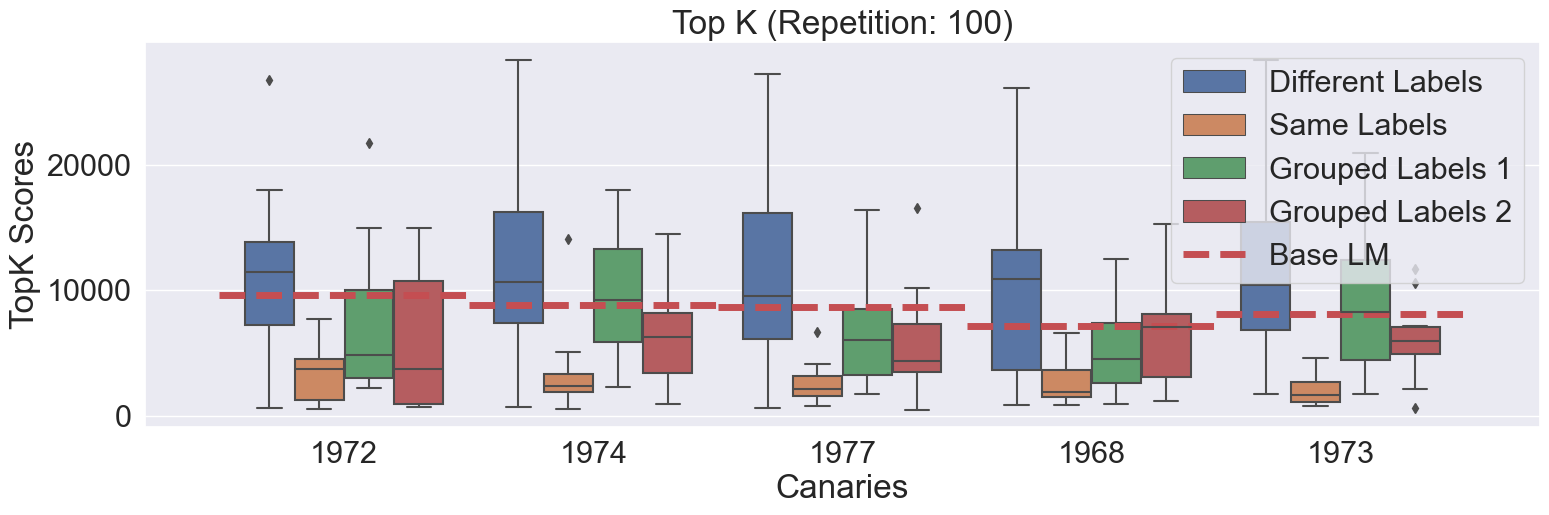}}
\subfigure[Exhaustive Search (rep=25).]{
\label{figure:multiCanaries_exhahustive_rep25}
\includegraphics[width=0.49\textwidth]{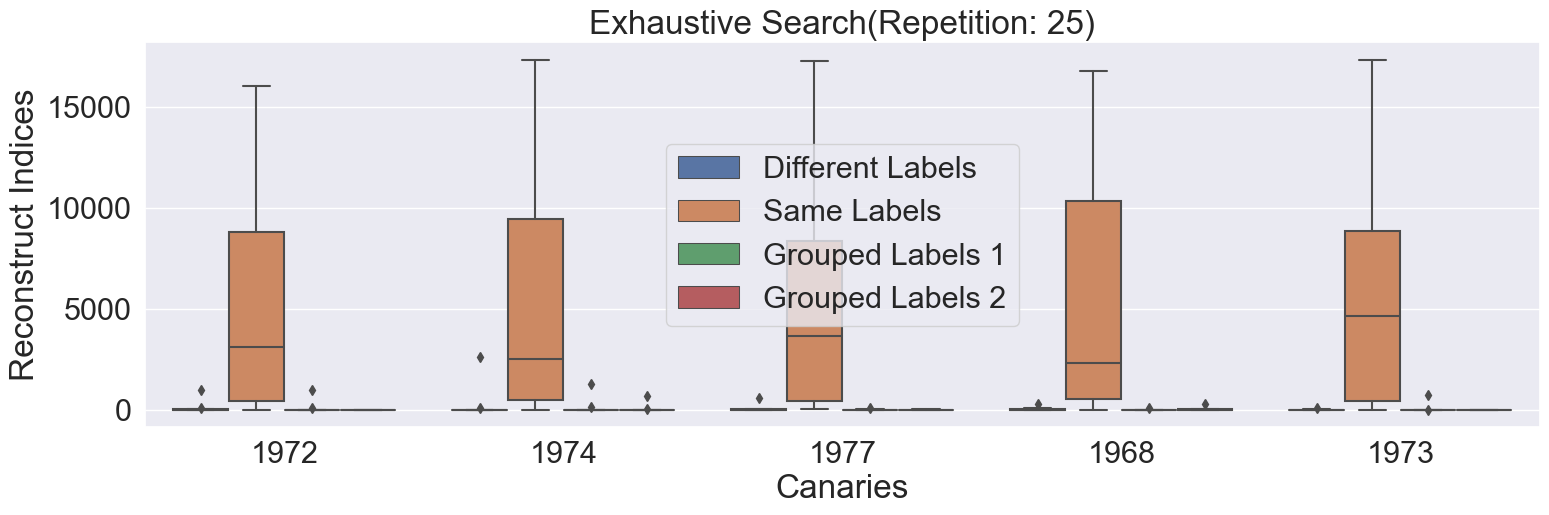}}
\subfigure[Exhaustive Search (rep=50).]{
\label{figure:multiCanaries_exhahustive_rep50}
\includegraphics[width=0.49\textwidth]{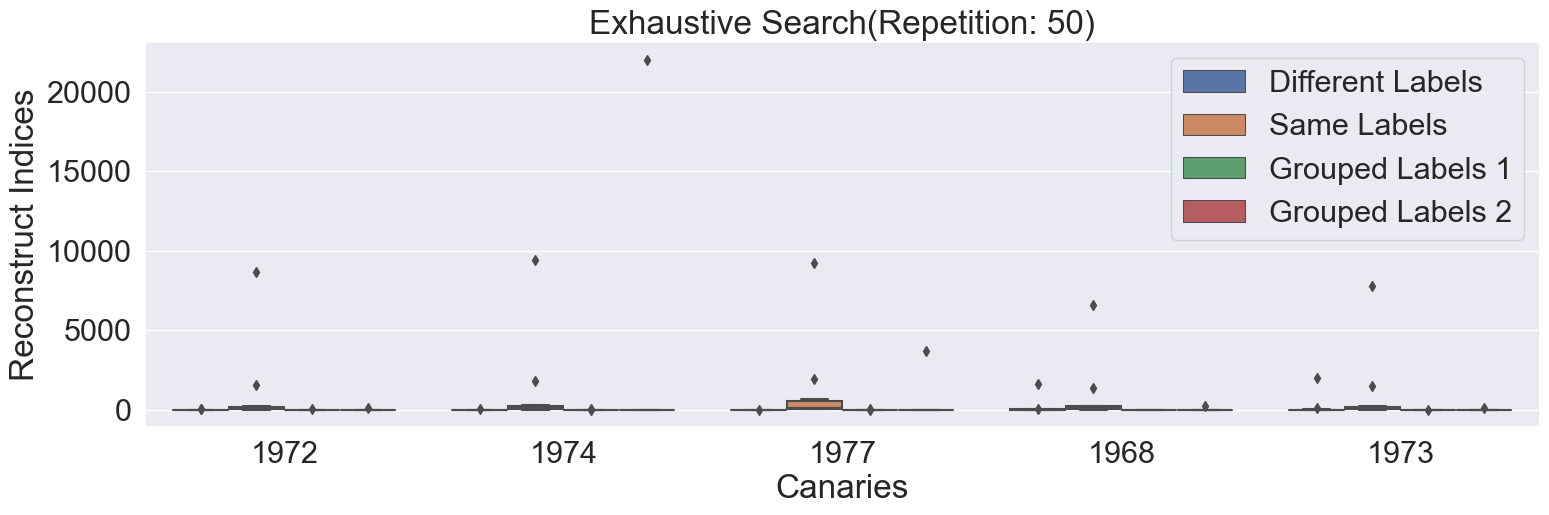}}
\subfigure[Exhaustive Search (rep=100).]{
\label{figure:multiCanaries_exhahustive_rep100}
\includegraphics[width=0.49\textwidth]{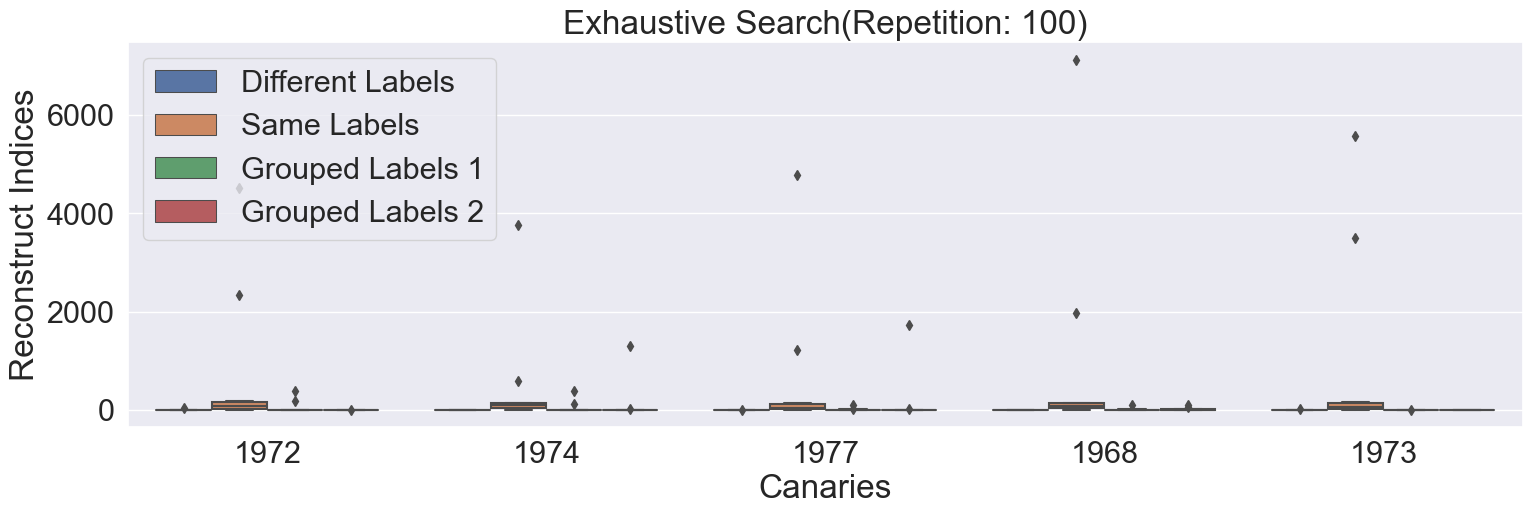}}
\subfigure[Language Model (rep=25).]{
\label{figure:multiCanaries_langModel_rep25}
\includegraphics[width=0.49\textwidth]{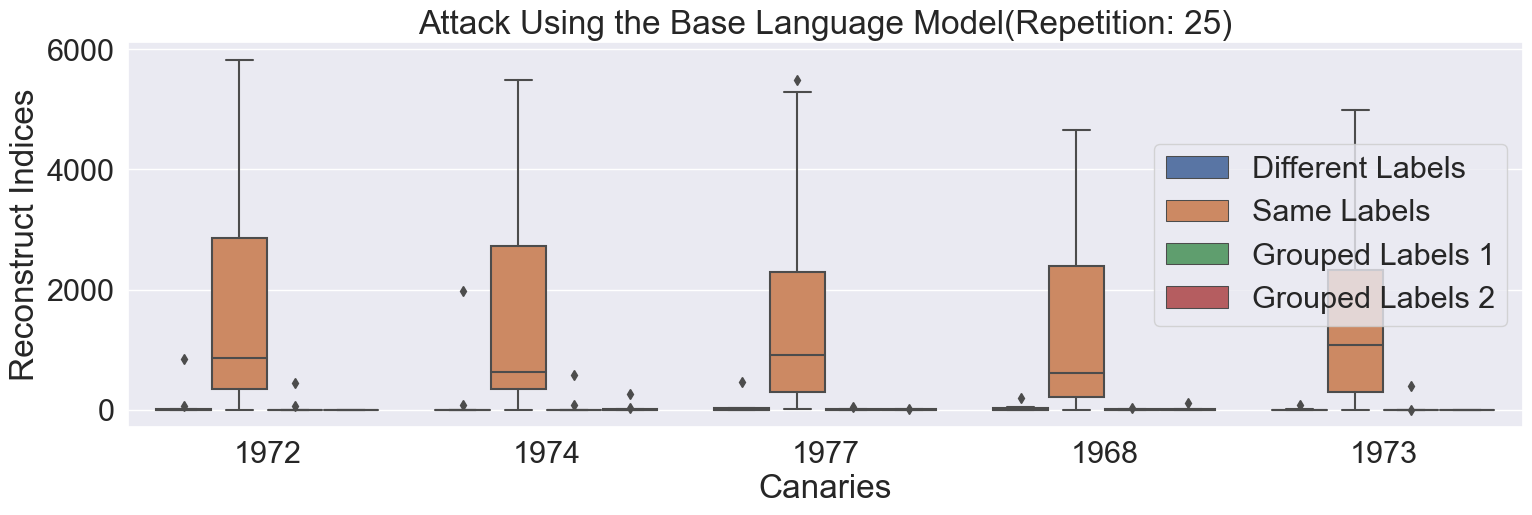}}
\subfigure[Language Model (rep=50).]{
\label{figure:multiCanaries_langModel_rep50}
\includegraphics[width=0.49\textwidth]{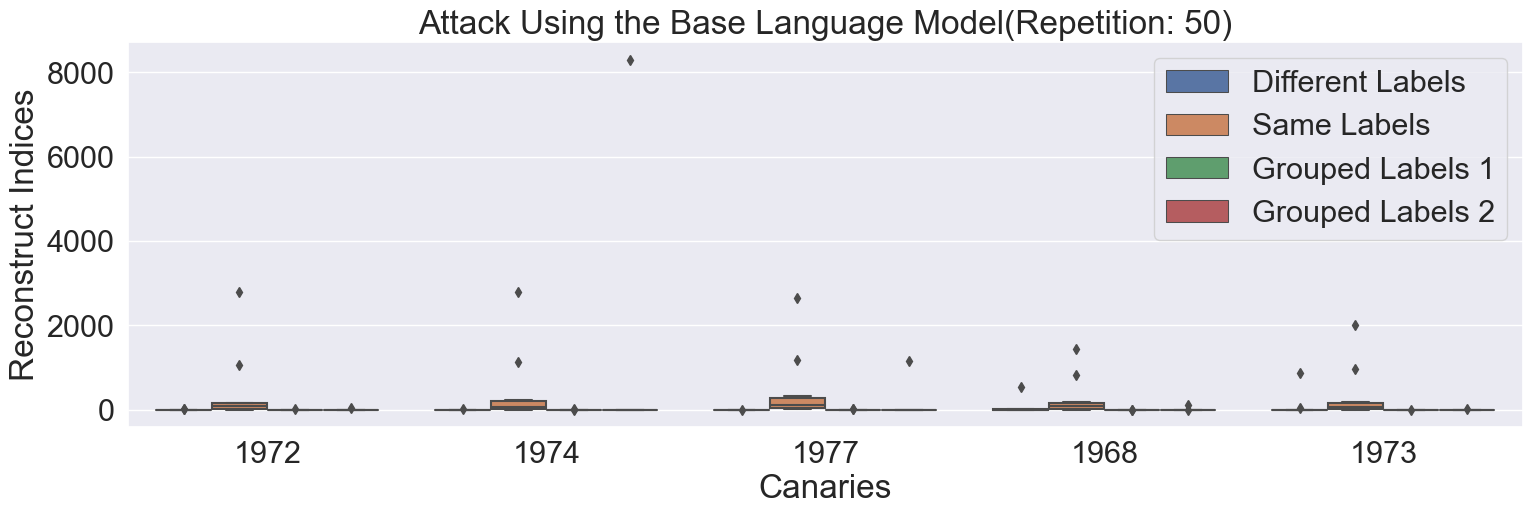}}
\subfigure[Language Model (rep=100).]{
\label{figure:multiCanaries_langModel_rep100}
\includegraphics[width=0.49\textwidth]{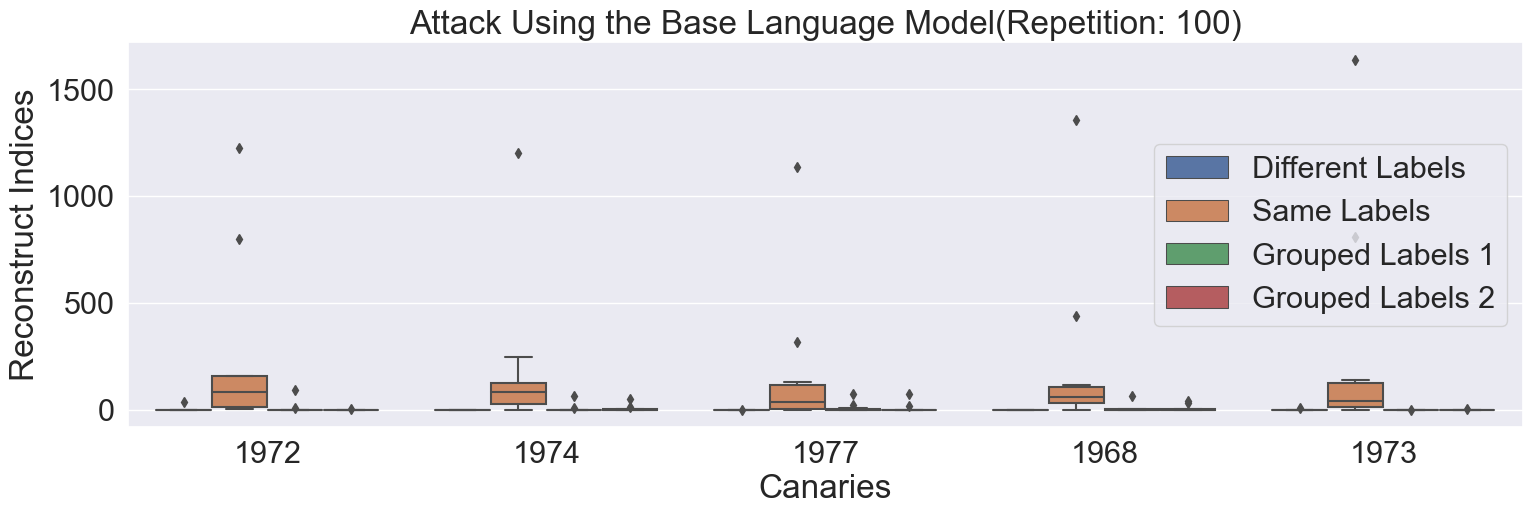}}
\subfigure[Frankenstein Model (rep=25).]{
\label{figure:multiCanaries_frankenstein_rep25}
\includegraphics[width=0.49\textwidth]{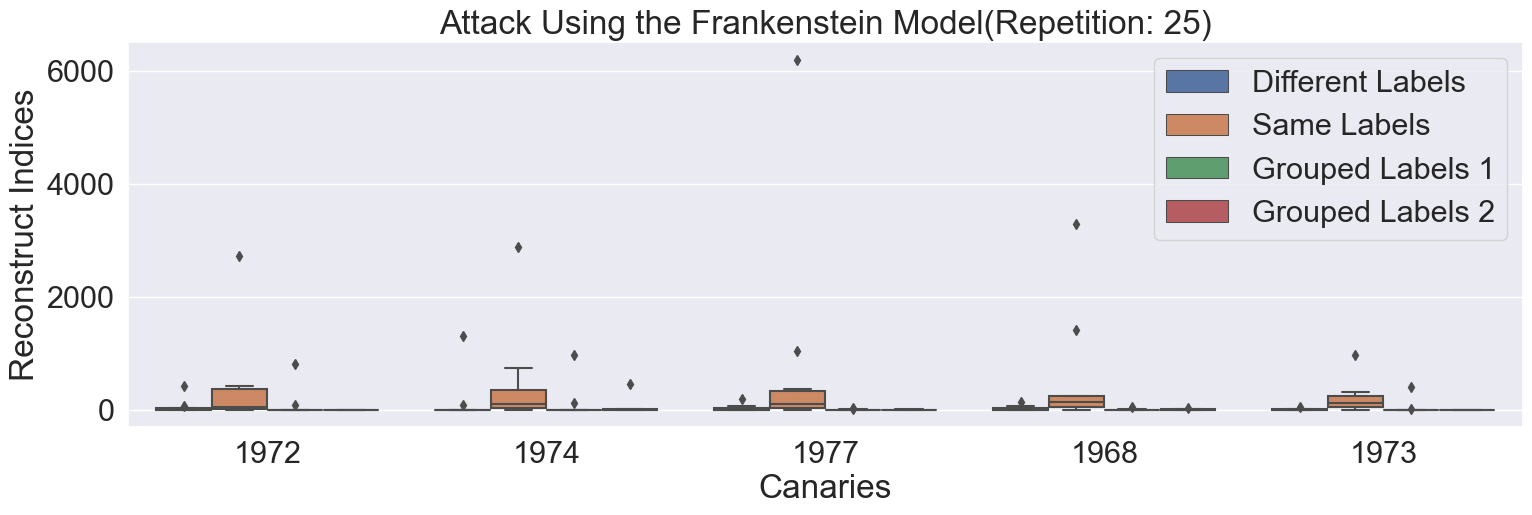}}
\subfigure[Frankenstein Model (rep=50).]{
\label{figure:multiCanaries_frankenstein_rep50}
\includegraphics[width=0.49\textwidth]{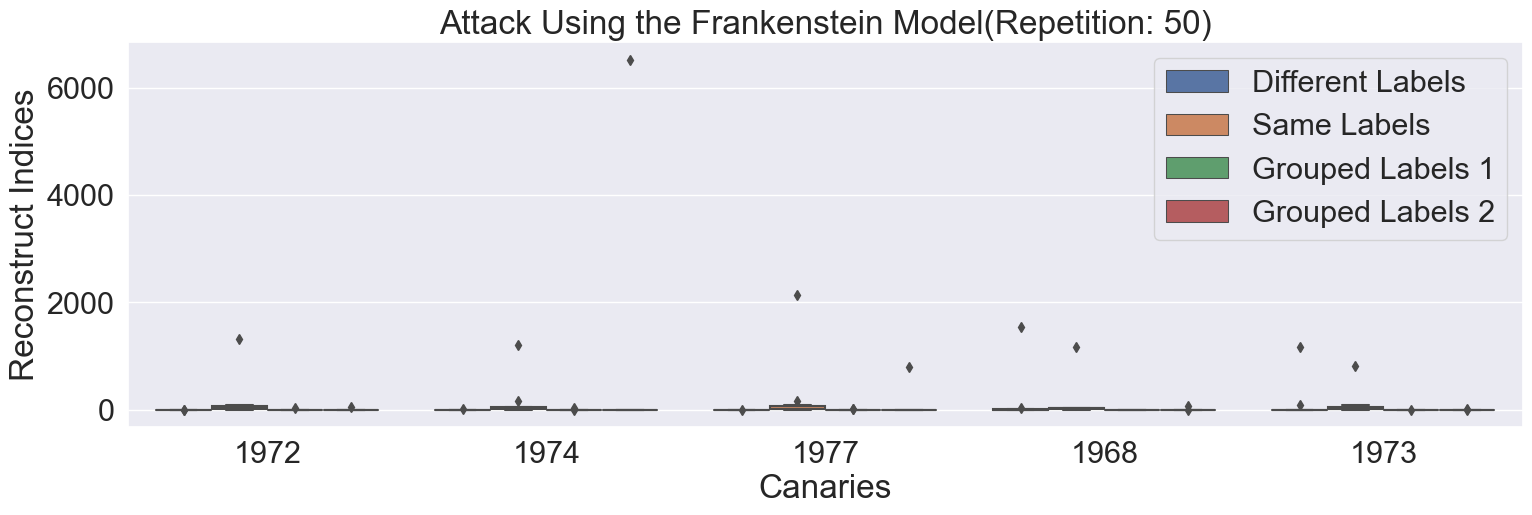}}
\subfigure[Frankenstein Model (rep=100).]{
\label{figure:multiCanaries_frankenstein_rep100}
\includegraphics[width=0.49\textwidth]{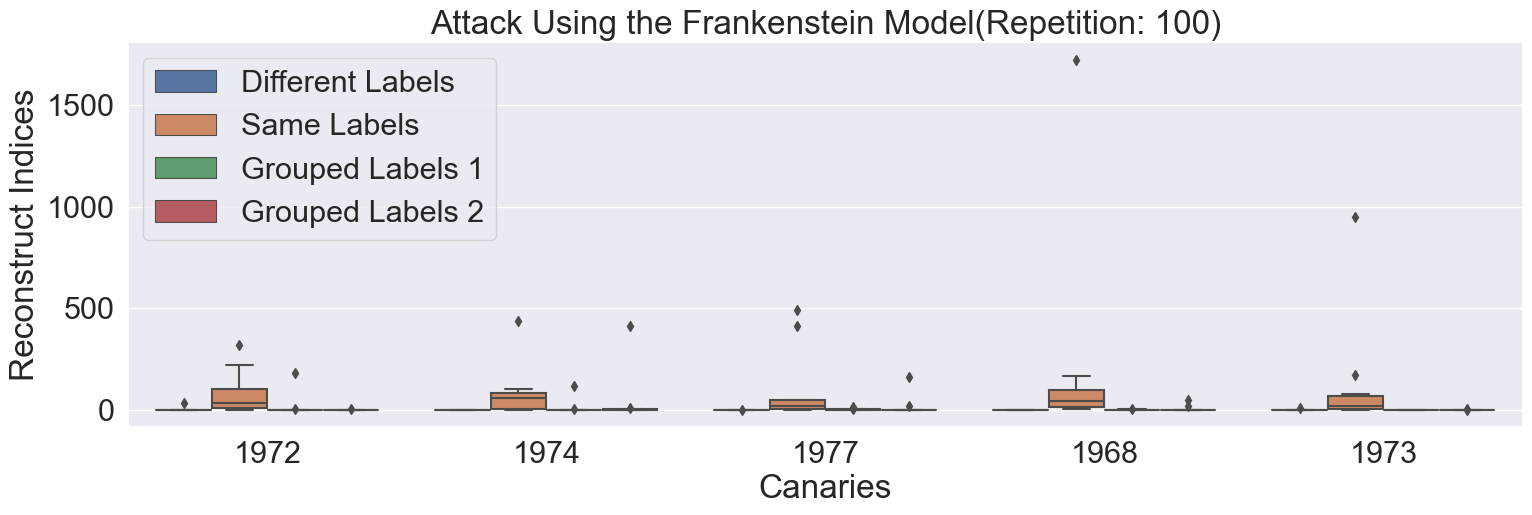}}
\caption{Effect of having multiple canaries with distinct class label patterns, varying only in the last token on the attack reconstruction on the Yelp reviews dataset.}
\label{figure:multiCanaries_yelp_exp}
\end{figure}

To assess the impact of canary labeling, we use the template "My social security number is [MASK]" and substitute the mask with five different values ("1972", "1974", "1977", "1968", "1973"). We then assign these sentences with either the same label, different labels for each sentence, or a combination of shared labels. Next, we inserted these sentences into the training data separately for each case. Fig.~\ref{figure:multiCanaries_yelp_exp} corroborates our expectation that utilizing canaries with distinct labels and differing by a single token greatly enhances the performance of the reconstruction attack. This finding highlights the potential risks of adversarial manipulation, where adversaries intentionally poison the training data by mislabeling specially constructed inputs to bolster the model's effectiveness against specific inputs.

\begin{figure*}[!t]
\centering
\subfigure[Top-K.]{
\label{figure:tokPos_topk}
\includegraphics[width=0.8\textwidth]{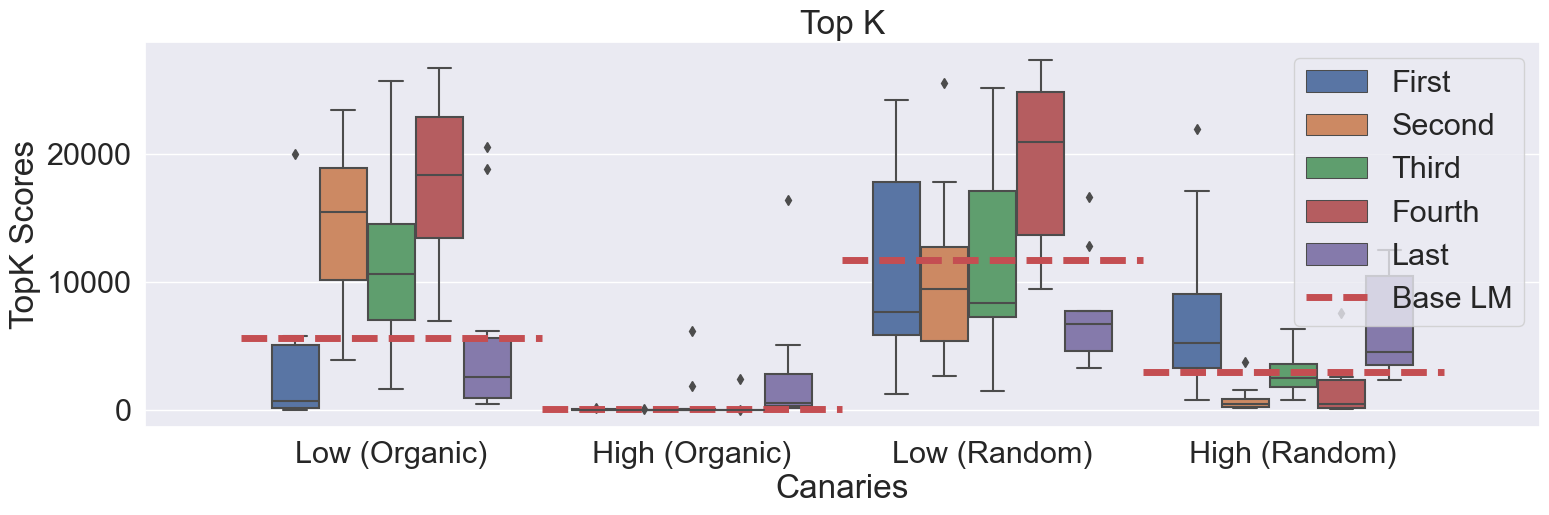}}
\subfigure[Exhaustive Search.]{
\label{figure:tokPos_exhaustive}
\includegraphics[width=0.8\textwidth]{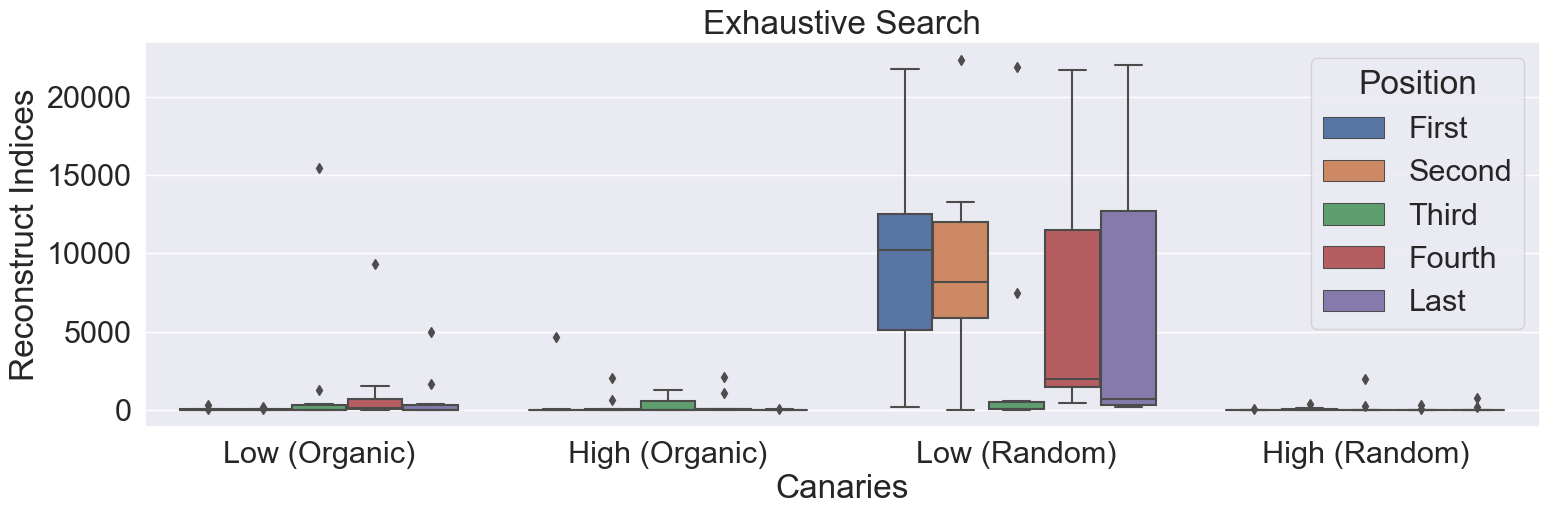}}
\subfigure[Language Model.]{
\label{figure:tokPos_langModel}
\includegraphics[width=0.8\textwidth]{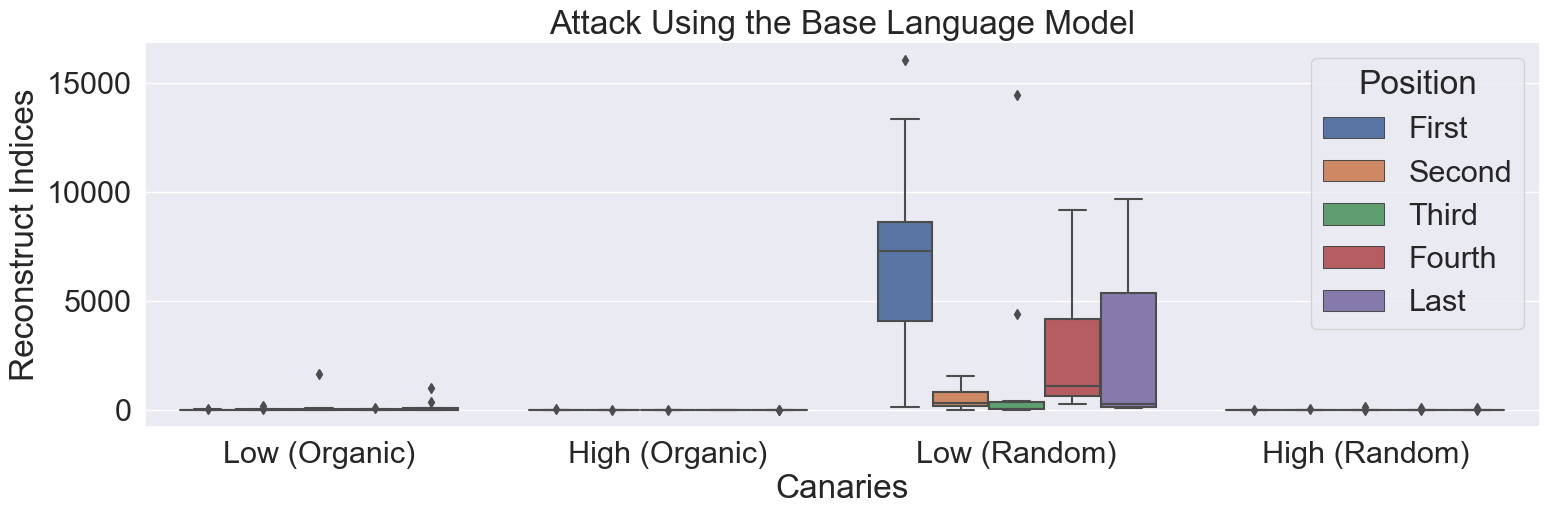}}
\subfigure[Frankenstein Model.]{
\label{figure:tokPos_frankenstein}
\includegraphics[width=0.8\textwidth]{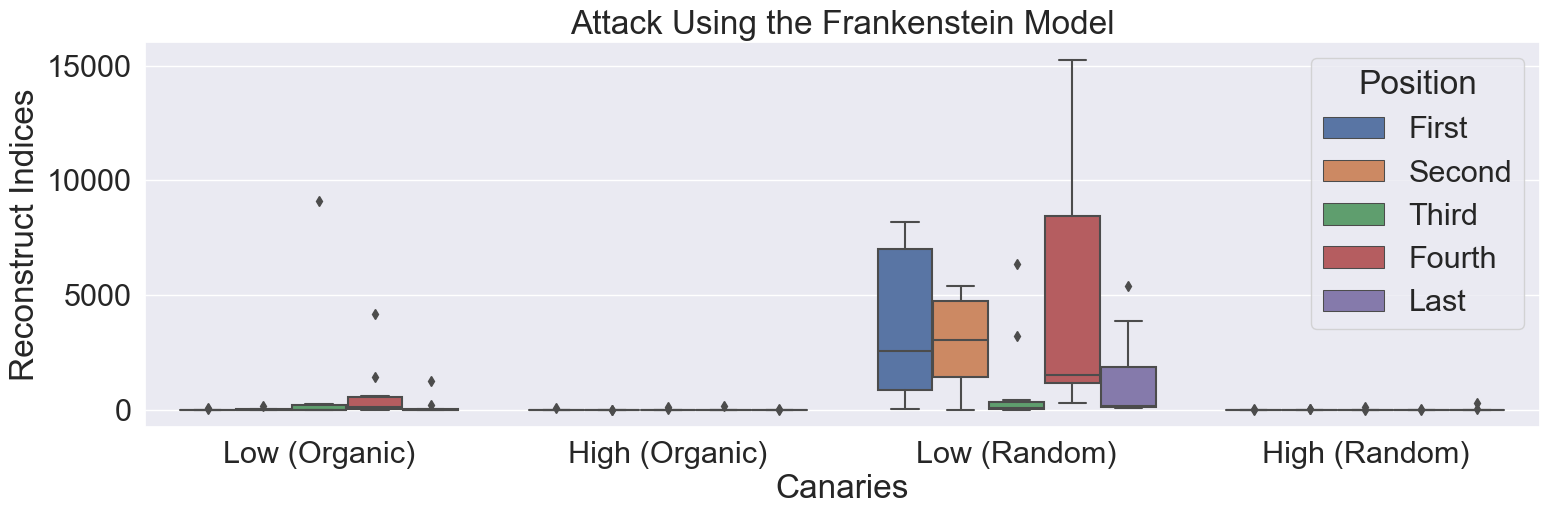}}
\caption{Effect of the position of the reconstructed token on the attack reconstruction on the Yelp reviews dataset under the same underlying model.}
\label{figure:tokPos_yelp}
\end{figure*}

Next, we examine the impact of different positions within the canaries. To that end, we analyze each token in four distinct canaries, each consisting of five tokens. Across all canaries, a consistent pattern was not discernible from our findings depicted in Fig.~\ref{figure:tokPos_yelp}. This lack of consistency can be attributed to variations in canary construction, such as their organic or random nature and the frequency of tokens used. For example, in organic canaries constructed from low-frequency tokens, the first and last positions yielded the best reconstruction performance, while the opposite was true for canaries constructed randomly from high-frequency tokens, where the first and last positions had the worst performance.

\begin{figure*}[!t]
\centering
\subfigure[Top-K.]{
\label{figure:canaryLen_topk}
\includegraphics[width=0.8\textwidth]{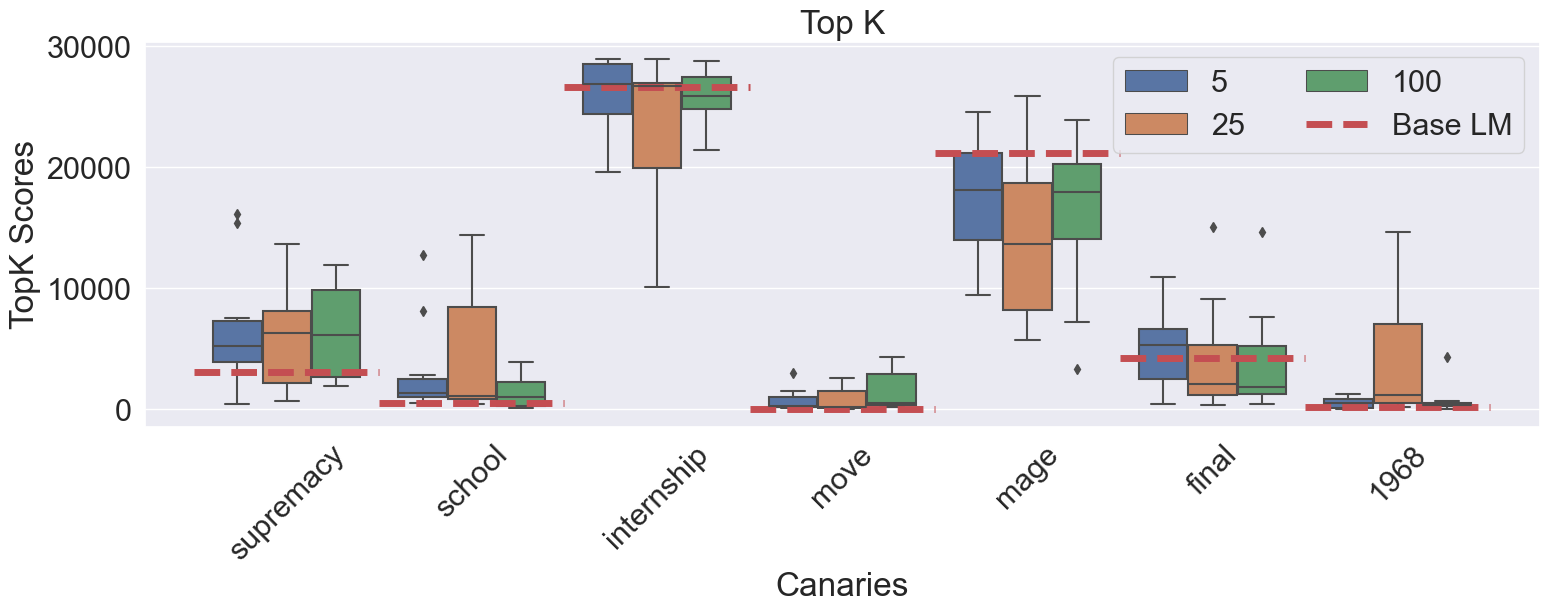}}
\subfigure[Exhaustive Search.]{
\label{figure:canaryLen_exhaustive}
\includegraphics[width=0.8\textwidth]{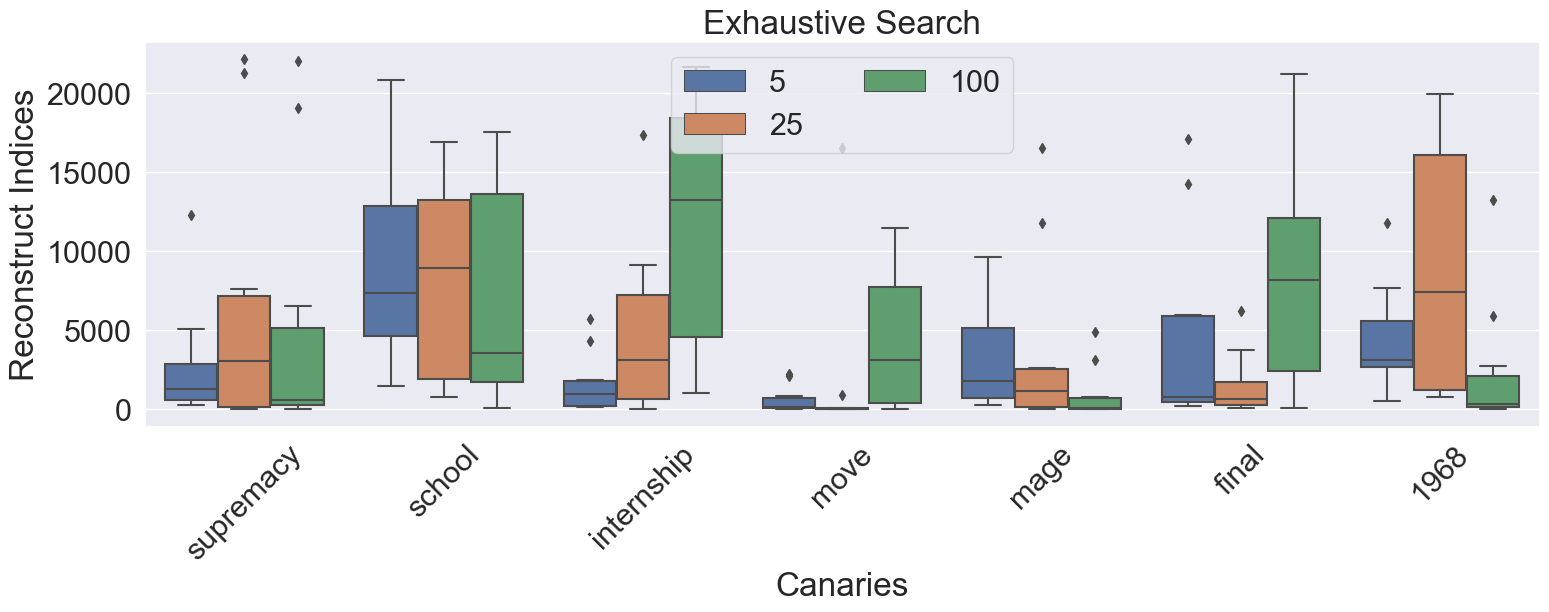}}
\subfigure[Language Model.]{
\label{figure:canaryLen_langModel}
\includegraphics[width=0.8\textwidth]{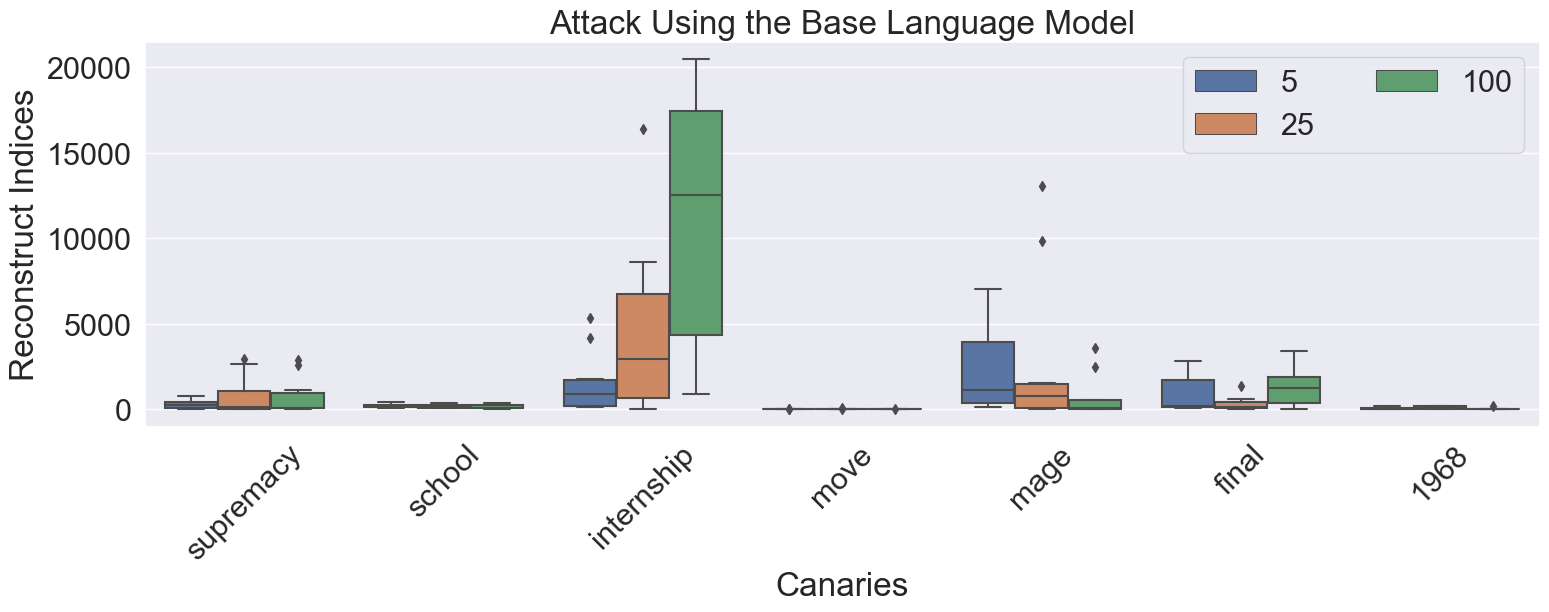}}
\subfigure[Frankenstein Model.]{
\label{figure:canaryLen_frankenstein}
\includegraphics[width=0.8\textwidth]{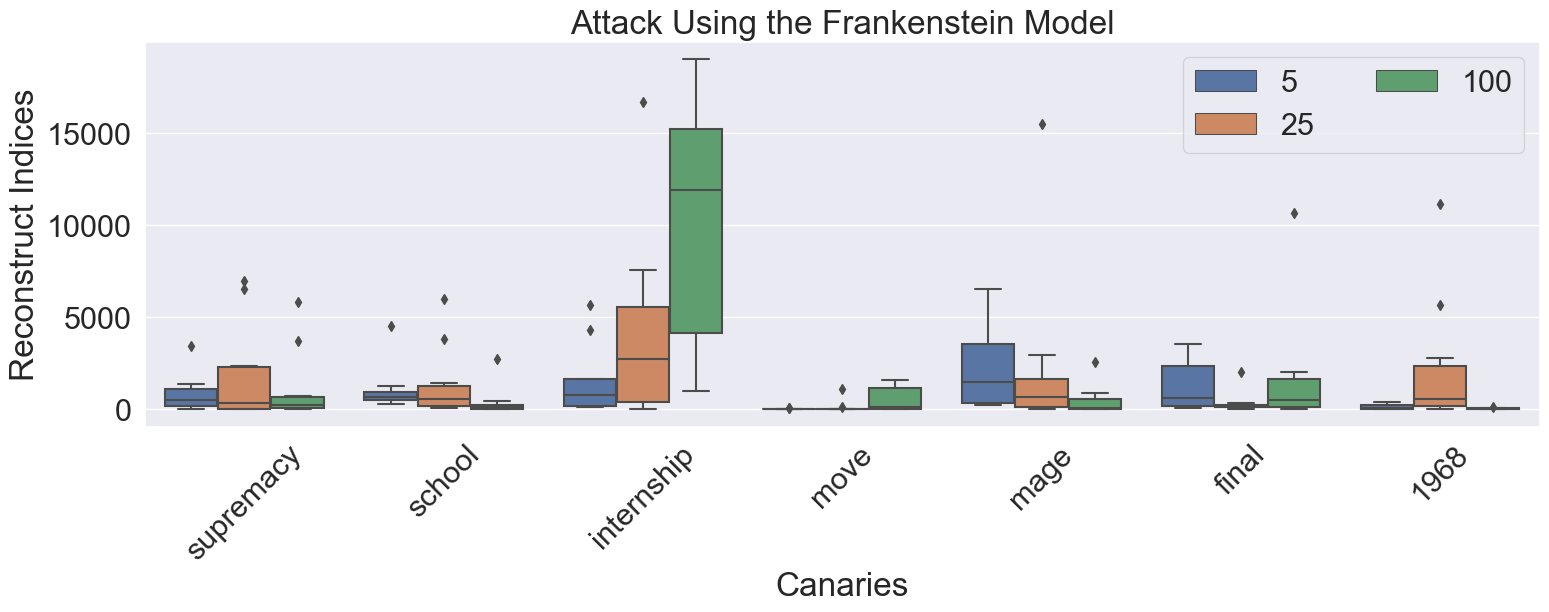}}
\caption{Effect of the canary length on the attack reconstruction on the Yelp reviews dataset for different repetition numbers of the canary.}
\label{figure:canaryLen_yelp}
\end{figure*}

Lastly, we assess the impact of increasing the size of the canaries by combining pairs of canaries from the same category using the ``and'' token. The reconstruction attacks are performed to construct the last token before the ending dot (``.''). When we compare the results presented in Fig.~\ref{figure:canaryLen_yelp} to those obtained when the canaries were roughly half the size (as shown in Fig.~\ref{fig:yelp_exp}), we observe that the performance remains relatively unchanged.

\section{Discussion And Limitation}
\subsection{\textbf{Limitations}}
The results of the experiments demonstrated the risks posed by data reconstruction attacks against classification models. However, we must acknowledge the limitations of our current attack methodology. The primary constraint lies in the number of target tokens that can be considered. Although increasing the number of target tokens introduces more uncertainty, our attack still outperforms the baseline. Nonetheless, we believe that future research can refine our attack approach to achieve better reconstruction of multiple target tokens. Additionally, it is important to note that our attack applies only to classification models that are fine-tuned on top of an LLM. Nevertheless, this setting is widely adopted in current practices, and we can leverage a public language model to generate candidate tokens without any alterations to the remaining steps.

\subsection{\textbf{Broader Impact}}
The focus of this study is to examine the potential privacy concerns arising from training a text classification model on sensitive and private data and to determine if any data leakage could occur in such a setting. This research serves as an initial investigation into the vulnerability of the text classification model to privacy breaches and identifying any misuse of personal data. It is worth noting that both the dataset and model used in this study are available to the public. 

\subsection{\textbf{Discussion}}
Our attack paves the way for various extensions and future research avenues. For instance, one possibility is to apply the attack on non-masked LLM, such as GPT-based models. By leveraging these models, adversaries can execute more intricate attacks by generating a substantial amount of text and subsequently pruning it, rather than focusing solely on individual target tokens.  Another approach is to explore the incorporation of an intermediate layer, such as an adapter, to enhance the connectivity between the generation head and the base model in the construction of the Frankenstein model. Alternatively, the adversary can explore the recent advancements in prompt-based learning to optimize a prompt that facilitates the connection between the base model and the generation head, thereby generating more effective candidate tokens.

\subsection{\textbf{Defense}}
Our attack consists of two phases, namely candidate generation and candidate pruning. Therefore, successfully defending against either of these phases would effectively defend against the attack as a whole. Since the candidate pruning phase heavily relies on the membership inference attack, defending against membership inference would successfully counter the \attack. One proven defense approach is to implement differential privacy with an appropriate privacy budget ($\epsilon$), which is guaranteed to provide defense against our attack. However, it is important to note that this defense mechanism may come at the expense of reduced utility.

\section{Conclusion}
This study represents the first comprehensive investigation of the reconstruction attack, shedding light on the crucial role of canary construction in determining the attack's outcomes. Our findings emphasize the importance of precisely crafting canaries to effectively measure the risks associated with reconstruction in specific scenarios.

\clearpage
\begin{table}[t!]
\caption{Data reconstruction attack on the Yelp and Reddit datasets with the canary being repeated 5 times. The reported values of top $K$ scores and beam sizes are obtained by averaging across a set of 10 runs, where each run uses different random seeds.}
\centering
\resizebox{\columnwidth}{!}{
\begin{tabular}{c|c@{\hspace{6pt}}c@{\hspace{6pt}}c@{\hspace{6pt}}c@{\hspace{6pt}}c@{\hspace{6pt}}c|c@{\hspace{6pt}}c@{\hspace{6pt}}c@{\hspace{6pt}}c@{\hspace{6pt}}c@{\hspace{6pt}}c}
\toprule
\multirow{2}{*}{} & \multicolumn{6}{c}{Yelp Dataset} & \multicolumn{6}{c}{Reddit Dataset} \\
\cmidrule(lr){2-7} \cmidrule(lr){8-13}
\multirow{2}{*}{} & \multicolumn{2}{c}{Exhaustive Search} & \multicolumn{2}{c}{Language Model} & \multicolumn{2}{c}{Frankenstein Model} & \multicolumn{2}{c}{Exhaustive Search} & \multicolumn{2}{c}{Language Model} & \multicolumn{2}{c}{Frankenstein Model} \\
\cmidrule(lr){2-7} \cmidrule(lr){8-13}
Target Token & Top K & Beam Size & Top K & Beam Size & Top K & Beam Size & Top K & Beam Size & Top K & Beam Size & Top K & Beam Size \\
\midrule
towels     & 22413 & 3390   &  3066  &  400    & 11856   & 1098    & 22413 & 3394.0   &    3066& 372.0      & 9664.0    & 1520.0    \\
supremacy  & 22413 & 2350   &  2327  &   251    & 2536    & 153    & 22413 & 2570.0   &    2327& 286.0      & 2989.0    & 354.0    \\
film       & 22413 & 5864   &   4551 &   1183    & 8181    & 2424  & 22413 & 6481.0   &  4551  & 1368.0      & 6154.0    & 2703.0   \\
school     & 22413 & 2518   &   1258 &   148    & 2351    & 275    & 22413 & 4366.0   &   1258 & 252.0      & 6828.0    & 1065.0    \\
flatly     & 22413 & 1638   &  11128  &  729     & 5850    & 340   & 22413 & 8387.0   &   11128 &3944.0       & 9300.0    & 2738.0   \\
internship & 22413 & 2831   &  25646  &   2590    & 19193   & 1311 & 22413 & 3882.0   & 25646   &3530.0       & 23785.0   & 3353.0   \\
best       & 22413 & 451    &  128  &   1    & 713    & 13         & 22413 & 1884.0    & 128   & 9.0      & 445.0    & 19.0     \\
move       & 22413 & 4157  &  10  &  1     & 289     & 65          & 22413 & 2534.0   &  10  & 2.0      & 892.0     & 88.0     \\
Fusion     & 22413 & 536   &  14541  &  249     & 18750   & 391    & 22413 & 3096.0   & 14541   & 1772.0      & 16664.0   & 2363.0   \\
mage       & 22413 & 634    &  11049  &   363    & 7283    & 417   & 22413 & 155.0    & 11049   & 70.0      & 8320.0   & 40.0     \\
night      & 22413 & 1716   &  1717  &  108     & 4496    & 475    & 22413 & 908.0   &  1717  &   41.0    & 5579.0    & 221.0    \\
final      & 22413 & 2304    &  4595  &  379     & 6005    & 1121  & 22413 & 1321.0   &4595    & 255.0      & 4192.0    & 220.0    \\
1974       & 22413 & 2861   &  8819  &   1064    & 5735    & 547   & 22413 & 5738.0   & 8819   & 1913.0      & 3771.0    & 847.0    \\
1968       & 22413 & 2601   &  7156  &  563     & 9795    & 1474   & 22413 & 5951.0   & 7156   & 1893.0      & 3810.0    & 556.0    \\
\bottomrule
\end{tabular}
}
\label{table:comparison_yelp_reddit_rep5}
\end{table}

\begin{table}[h!]
\caption{Data reconstruction attack on the Yelp and Reddit datasets with the canary being repeated 25 times. The reported values of top $K$ scores and beam sizes are obtained by averaging across a set of 10 runs, where each run uses different random seeds.}
\centering
\resizebox{\columnwidth}{!}{
\begin{tabular}{c|c@{\hspace{6pt}}c@{\hspace{6pt}}c@{\hspace{6pt}}c@{\hspace{6pt}}c@{\hspace{6pt}}c|c@{\hspace{6pt}}c@{\hspace{6pt}}c@{\hspace{6pt}}c@{\hspace{6pt}}c@{\hspace{6pt}}c}
\toprule
\multirow{2}{*}{} & \multicolumn{6}{c}{Yelp Dataset} & \multicolumn{6}{c}{Reddit Dataset} \\
\cmidrule(lr){2-7} \cmidrule(lr){8-13}
\multirow{2}{*}{} & \multicolumn{2}{c}{Exhaustive Search} & \multicolumn{2}{c}{Language Model} & \multicolumn{2}{c}{Frankenstein Model} & \multicolumn{2}{c}{Exhaustive Search} & \multicolumn{2}{c}{Language Model} & \multicolumn{2}{c}{Frankenstein Model} \\
\cmidrule(lr){2-7} \cmidrule(lr){8-13}
Target Token & Top K & Beam Size & Top K & Beam Size & Top K & Beam Size & Top K & Beam Size & Top K & Beam Size & Top K & Beam Size \\
\midrule
towels     & 22413 & 5031.0   &  3066  &  740.0     & 12820.0   & 2681.0   & 22413 & 2753.0   &3066    &352.0       & 9173.0    & 609.0    \\
supremacy  & 22413 & 2943.0   & 2327   &   288.0    & 4666.0    & 1145.0   & 22413 & 1392.0   &2327    &291.0       & 3677.0    & 195.0    \\
film       & 22413 & 5808.0   &  4551  &   1109.0    & 5539.0    & 1414.0  & 22413 & 3556.0   &4551    &608.0       & 7164.0    & 986.0   \\
school     & 22413 & 856.0   &  1258  &   28.0    & 4753.0    & 133.0      & 22413 & 1371.0   &1258    &69.0       & 2462.0    & 267.0    \\
flatly     & 22413 & 2170.0   &  11128  & 946.0      & 10208.0    & 1501.0 & 22413 & 1349.0   &11128    &681.0       & 6574.0    & 441.0   \\
internship & 22413 & 2354.0   &  25646  & 2084.0      & 13297.0   & 1145.0 & 22413 & 2777.0   &25646    &2516.0       & 17293.0   & 1949.0   \\
best       & 22413 & 73.0    &  128  &  1.0     & 1305.0    & 24.0         & 22413 & 240.0    &128    & 2.0      & 1930.0    & 5.0     \\
move       & 22413 & 232.0   &  10  &  0.0     & 550.0     & 9.0           & 22413 & 472.0   &10    & 0.0      & 774.0     & 10.0     \\
Fusion     & 22413 & 1896.0   &  14541  &  941.0     & 22533.0   & 1446.0  & 22413 & 988.0   &14541    &461.0       & 17817.0   & 639.0   \\
mage       & 22413 & 1581.0    &  11049  & 677.0      & 11013.0    & 806.0 & 22413 & 305.0    &11049    & 148.0      & 13662.0   & 84.0     \\
night      & 22413 & 4937.0   &  1717  &   298.0    & 3484.0    & 395.0    & 22413 & 1277.0   &1717    &101.0       & 7155.0    & 277.0    \\
final      & 22413 & 563.0    &  4595  &  107.0     & 4034.0    & 68.0     & 22413 & 177.0   &4595    &28.0       & 3188.0    & 19.0    \\
1974       & 22413 & 2639.0   &  8819  &  777.0     & 7540.0    & 528.0    & 22413 & 260.0   &8819    & 98.0      & 5944.0    & 102.0    \\
1968       & 22413 & 1607.0   &  7156  &  390.0     & 7392.0    & 793.0    & 22413 & 308.0   &7156    & 98.0      & 6212.0    & 210.0    \\
\bottomrule
\end{tabular}
}
\label{table:comparison_yelp_reddit_rep25}
\end{table}

\begin{table}[h!]
\caption{Data reconstruction attack on the Yelp and Reddit datasets with the canary being repeated 100 times. The reported values of top $K$ scores and beam sizes are obtained by averaging across a set of 10 runs, where each run uses different random seeds.}
\centering
\resizebox{\columnwidth}{!}{
\begin{tabular}{c|c@{\hspace{6pt}}c@{\hspace{6pt}}c@{\hspace{6pt}}c@{\hspace{6pt}}c@{\hspace{6pt}}c|c@{\hspace{6pt}}c@{\hspace{6pt}}c@{\hspace{6pt}}c@{\hspace{6pt}}c@{\hspace{6pt}}c}
\toprule
\multirow{2}{*}{} & \multicolumn{6}{c}{Yelp Dataset} & \multicolumn{6}{c}{Reddit Dataset} \\
\cmidrule(lr){2-7} \cmidrule(lr){8-13}
\multirow{2}{*}{} & \multicolumn{2}{c}{Exhaustive Search} & \multicolumn{2}{c}{Language Model} & \multicolumn{2}{c}{Frankenstein Model} & \multicolumn{2}{c}{Exhaustive Search} & \multicolumn{2}{c}{Language Model} & \multicolumn{2}{c}{Frankenstein Model} \\
\cmidrule(lr){2-7} \cmidrule(lr){8-13}
Target Token & Top K & Beam Size & Top K & Beam Size & Top K & Beam Size & Top K & Beam Size & Top K & Beam Size & Top K & Beam Size \\
\midrule
towels     & 22413 & 7137.0   & 3066   &  1016.0     & 11970.0   & 3150.0   & 22413 & 905.0   &3066    &  99.0     & 8114.0    & 238.0    \\
supremacy  & 22413 & 5984.0   & 2327   &  660.0     & 2428.0    & 610.0     & 22413 & 1626.0   &2327    & 214.0      & 4656.0    & 131.0    \\
film       & 22413 & 3052.0   & 4551   &  508.0     & 6226.0    & 881.0     & 22413 & 170.0   &4551    &  27.0     & 3472.0    & 22.0   \\
school     & 22413 & 6608.0   & 1258   &  314.0     & 1705.0    & 331.0     & 22413 & 549.0   &1258    &  28.0     & 6188.0    & 190.0    \\
flatly     & 22413 & 6798.0   & 11128   & 2856.0      & 9434.0    & 1646.0  & 22413 & 258.0   &11128    & 63.0      & 7111.0    & 25.0   \\
internship & 22413 & 5516.0   & 25646   & 5004.0      & 18940.0   & 4697.0  & 22413 & 3743.0   &25646    &3366.0       & 15999.0   & 2724.0   \\
best       & 22413 & 7.0    & 128   &    0.0   & 5099.0    & 4.0            & 22413 & 99.0    &128    & 1.0      & 1986.0    & 24.0     \\
move       & 22413 & 1149.0   & 10   &    0.0   & 498.0     & 37.0          & 22413 & 2.0   &10    & 0.0      & 1153.0     & 0.0     \\
Fusion     & 22413 & 2854.0   & 14541   & 1539.0      & 21566.0   & 2240.0  & 22413 & 862.0   &14541    & 453.0      & 18852.0   & 546.0   \\
mage       & 22413 & 362.0    & 11049   & 157.0      & 7884.0    & 38.0     & 22413 & 144.0    &11049    &89.0       & 11630.0   & 110.0     \\
night      & 22413 & 4013.0   & 1717   &  273.0     & 3074.0    & 377.0     & 22413 & 2107.0   &1717    & 203.0      & 3656.0    & 326.0    \\
final      & 22413 & 7.0    & 4595   &    1.0   & 3669.0    & 2.0           & 22413 & 1625.0   & 4595   & 259.0      & 3949.0    & 346.0    \\
1974       & 22413 & 281.0   & 8819   &   169.0    & 8860.0    & 158.0      & 22413 & 265.0   &8819    & 116.0      & 4555.0    & 48.0    \\
1968       & 22413 & 202.0   & 7156   &   71.0    & 2394.0    & 16.0        & 22413 & 65.0   &7156    & 45.0      & 3200.0    & 28.0    \\
\bottomrule
\end{tabular}
}
\label{table:comparison_yelp_reddit_rep100}
\end{table}

\clearpage
\bibliographystyle{IEEEtran}
\bibliography{refs.bib}

\end{document}